\documentclass[letterpaper]{article} 
\usepackage{aaai24}  
\usepackage{times}  
\usepackage{helvet}  
\usepackage{courier}  
\usepackage[hyphens]{url}  
\usepackage{graphicx} 
\usepackage{natbib}  
\usepackage{caption} 
\usepackage{algorithm}
\usepackage{algorithmic}

\usepackage{amsmath,amsfonts}

\usepackage{amssymb}
\usepackage{subcaption}
\usepackage{tabularx}
\usepackage{multirow}
\usepackage{amsthm}
\usepackage{bm}  
\usepackage{tikz}
\usepackage{makecell}
\usepackage{dsfont}

\newcommand{\X}{\mathbf{X}}
\newcommand{\D}{\mathbf{D}}
\newcommand{\x}{\mathbf{x}}
\newcommand{\Y}{\mathbf{Y}}
\newcommand{\y}{\mathbf{y}}
\newcommand{\Z}{\mathbf{Z}}
\newcommand{\z}{\mathbf{z}}

\newcommand{\BoldU}{\mathbf{U}}
\newcommand{\W}{\mathbf{W}}
\newcommand{\w}{\mathbf{w}}

\newcommand{\BC}{\mathbf{C}}
\newcommand{\V}{\mathbf{V}}
\newcommand{\U}{\mathbf{U}}

\usepackage{color}

\DeclareMathOperator*{\argmax}{arg\,max}
\DeclareMathOperator*{\argmin}{arg\,min}
\newcommand\independent{\protect\mathpalette{\protect\independenT}{\perp}}
\def\independenT#1#2{\mathrel{\rlap{$#1#2$}\mkern2mu{#1#2}}}

\newcommand{\CG}{\mathcal{G}}

\newtheorem{theorem}{Theorem}
\newtheorem{lemma}{Lemma}
\newtheorem{definition}{Definition}

\newtheorem{theorem*}{Theorem}

\usepackage{newfloat}
\usepackage{listings}
\DeclareCaptionStyle{ruled}{labelfont=normalfont,labelsep=colon,strut=off} 
\floatstyle{ruled}
\newfloat{listing}{tb}{lst}{}
\floatname{listing}{Listing}
\title{Robustly Improving Bandit Algorithms with Confounded and Selection Biased Offline Data: A Causal Approach}
\author{
    Wen Huang,
    Xintao Wu
}
\affiliations{
  University of Arkansas\\ 
  \{wenhuang , xintaowu\}@uark.edu
}

\usepackage{bibentry}

\begin{document}

\maketitle

\begin{abstract}
This paper studies bandit problems where an agent has access
to offline data that might be utilized to potentially improve the
estimation of each arm’s reward distribution. A major obstacle in this setting is the existence of compound biases from
the observational data. Ignoring these biases and blindly fitting a model with the biased data could even negatively affect the online learning phase. In this work, we formulate this
problem from a causal perspective. First, we categorize the
biases into confounding bias and selection bias based on the
causal structure they imply. Next, we extract the causal bound
for each arm that is robust towards compound biases from
biased observational data. The derived bounds contain the
ground truth mean reward and can effectively guide the bandit agent to learn a nearly-optimal decision policy. We also
conduct regret analysis in both contextual and non-contextual
bandit settings and show that prior causal bounds could help
consistently reduce the asymptotic regret.
\end{abstract}

\section{Introduction}

The past decade has seen the rapid development of contextual bandit as a legit framework to model interactive decision-making scenarios, such as personalized recommendation \citep{li2010contextual}, online advertising \citep{tang2013automatic,avadhanula2021stochastic}, and anomaly detection \citep{ding2019interactive}. The key challenge in a contextual bandit problem is to select the most beneficial item (i.e. the corresponding arm or intervention) according to the observed context at each round. 
In practice it is common that the agent has additional access to logged data from various sources, which may provide some useful information. A key issue is how to accurately leverage offline data such that it can efficiently assist the online decision-making process. However, one inevitable problem is that there may exist compound biases in the offline dataset, probably due to the data collection process, the existence of unobserved variables, the policies implemented by the agent, and so on \cite{chen2023bias}. As a consequence, blindly fitting a model without considering those biases will lead to an inaccurate estimator of the reward distribution for each arm, ending up inducing a negative impact on the online learning phase.

To overcome this limitation and make good use of the offline data for online bandit learning, we formulate our framework from a causal inference perspective. Causal inference provides a family of methods to infer the effects of actions from a combination of data and qualitative assumptions about the underlying mechanism. Based on Pearl's structural causal model \citep{pearl2009causality} we can derive a truncated factorization formula that expresses the target causal quantity with probability distributions from the data. Appropriately adopting that prior knowledge on the reward distribution of each arm can accelerate the learning speed and achieve lower cumulative regret for online bandit algorithms.

Previous studies along this direction \citep{zhang2017transfer,sharma2020warm,tennenholtz2021bandits} only focused on one specific  bias and have not dealt with compound biases in the offline data. It was shown in \citep{bareinboim2014recovering} that biases could be classified into confounding bias and selection bias based on the causal structure they imply. Due to the orthogonality of confounding and selection bias, simply deconfounding and estimating causal effects in the presence of selection bias using observational data is in general impractical without further assumptions, such as strong graphical conditions \citep{correa2017causal} or the accessibility of external unbiasedly measured data \citep{bareinboim2015recovering}. In this paper, we address this limitation by non-parametrically bounding the target conditional causal effect when 
confounding and selection biases can not be mitigated simultaneously. We propose two novel strategies to extract prior causal bounds for the reward distribution of each arm and use them to effectively guide the bandit agent to learn a nearly-optimal decision policy. We demonstrate that our approach could further reduce cumulative regret and is resistant to different levels of compound biases in offline data.  

Our contributions can be summarized into three parts: 1) We derive causal bounds for conditional causal effects under confounding and selection biases based on c-component factorization and substitute intervention methods; 2) we propose a novel framework that leverages the prior causal bound obtained from biased offline data to guide the arm-picking process in bandit algorithms, thus robustly decreasing the exploration of sub-optimal arms and reducing the cumulative regret; and 3) we develop one contextual bandit algorithm (LinUCB-PCB) and one non-contextual bandit algorithm (UCB-PCB) that are enhanced with prior causal bounds. We theoretically show under mild conditions both bandit algorithms achieve lower regret than their non-causal counterparts. We also conduct an empirical evaluation to demonstrate the effectiveness of our method under the linear contextual bandit setting. 

\section{Background}
\label{sec:background}

Our work is based on Pearl's structural causal model \citep{pearl2009causality} which describes the causal mechanisms of a system as a set of structural equations.

\begin{definition}[Structural Causal Model (SCM) \citep{pearl2009causality}]\label{def:cm}
A causal model $\mathcal{M}$ is a triple $\mathcal{M} = \langle \mathbf{U},\mathbf{V},\mathbf{F} \rangle$ where
1) $\mathbf{U}$ is a set of hidden contextual variables that are determined by factors outside the model; 	
2) $\mathbf{V}$ is a set of observed variables that are determined by variables in $\mathbf{U}\cup\mathbf{V}$;
3) $\mathbf{F}$ is a set of equations mapping from $\mathbf{U}\times \mathbf{V}$ to $\mathbf{V}$. Specifically, for each $V\in \mathbf{V}$, there is an equation 
$v = f_{V}(Pa(V),\mathbf{u}_{V})$ where $Pa(V)$ is a realization of a set of observed variables called the parents of $V$, and $\mathbf{u}_{V}$ is a realization of a set of hidden variables.
\end{definition}


Quantitatively measuring causal effects is facilitated with the $do$-operator \citep{pearl2009causality}, which simulates the physical interventions that force some variables to take certain values. Formally, the intervention that sets the value of $X$ to $x$ is denoted by $do(x)$. In a SCM, intervention $do(x)$ is defined as the substitution of equation $x=f_{X}(Pa(X),\mathbf{u}_{X})$ with constant $X=x$.
The causal model $\mathcal{M}$ is associated with a causal graph $\CG = \langle \mathcal{V}, \mathcal{E} \rangle$.
Each node of $\mathcal{V}$ corresponds to a variable of $\mathbf{V}$ in $\mathcal{M}$.
Each edge in $\mathcal{E}$, denoted by a directed arrow $\rightarrow$, points from a node $X\in \mathbf{U} \cup \mathbf{V}$ to a different node $Y\in \mathbf{V}$ if $f_Y$ uses values of $X$ as input. 
The intervention that sets the value of a set of variables $\mathbf{X}$ to $\mathbf{x}$ is denoted by $do(\mathbf{X} = \mathbf{x})$. 
The post-intervention distribution of the outcome variables $P(\mathbf{y}|do(\mathbf{x}))$ can be computed by the truncated factorization formula \citep{pearl2009causality}, 
\begin{equation}
P(\mathbf{y}|do(\mathbf{x})) = \prod_{Y\in \mathbf{Y}}P(y|Pa(Y))\delta_{\mathbf{X}=\mathbf{x}},    
\end{equation}
where $\delta_{\mathbf{X}=\mathbf{x}}$ means assigning attributes in $\mathbf{X}$ involved in the term ahead with the corresponding values in $\mathbf{x}$. The post intervention distribution $P(\mathbf{y}|do(\mathbf{x}))$ is  identifiable if it can be uniquely computed from observational distributions $P(\V)$.

\noindent \textbf{Confounding Bias} occurs when there exist hidden variables that simultaneously determine exposure variables and the outcome variable. It is well known that, in the absence of hidden confounders,  all causal effects can be estimated consistently from non-experimental data. However, in the presence of hidden confounders, whether the desired causal quantity can be estimated depends on the locations of the unmeasured variables, the intervention set, and the outcome.  
To adjust for confounding bias, one common approach is to condition on a set of covariates that satisfy the backdoor criterion.
\cite{shpitser2012validity} further generalized the backdoor criterion to identify the causal effect $P(\mathbf{y}|do(\mathbf{x}))$ if all non-proper causal paths are blocked. 

\begin{definition}[Generalized Backdoor Criterion \cite{shpitser2012validity}] \label{GBC} 
A set of variables $\mathbf{\Z}$ satisfies the adjustment criterion relative to $(\X, \Y)$ in $\mathcal{G}$  if: (i) no element in $\Z$ is a descendant in $\mathcal{G}_{\overline {\X}}$ of any $W \notin \X$ lying on a proper causal path from $\X$ to $Y$. (ii) all non-causal paths in $\mathcal{G}$ from $\X$ to $Y$ are blocked by $\Z$.
\end{definition}
In Definition \ref{GBC} $\mathcal{G}_{\overline{\X}}$ denotes the graph resulting from removing all incoming edges to $\X$ in $\mathcal{G}$, and a causal path from a node in $\mathbf{X}$ to $\Y$ is called proper if it does not intersect $\mathbf{X}$ except at the starting point. The causal effect can thus be computed by controlling for a set of covariates $\mathbf{Z}$.
\begin{equation}
    P(\y|do(\mathbf{x}))  = \sum_{\Z} P(\y| \mathbf{x},\z) P(\z)
\label{eq: backdoor adjustment}
\end{equation}
\noindent \textbf{Sample Selection Bias} arises with a biased selection mechanism, e.g., choosing users based on a certain time or location. To accommodate for SCM framework, we introduce a node $S$ in a causal graph representing a binary indicator of entry into the observed data, and denote the causal graph augmented with selection node $S$ as $\mathcal{G}_s$. 
Generally speaking, the target distribution $P(\y|do(\x))$ is called s-recoverable if it can be computed from the available (biased) observational distributions $P(\V|S=1)$ in the augmented graph $\mathcal{G}_s$.
To recover causal effects in the presence of confounding and sample selection bias, \cite{correa2018generalized} studied the use of generalized adjustment criteria and introduced a sufficient and necessary condition for recovering causal effects from biased distributions.  

\begin{theorem}[Generalized Adjustment for Causal Effect \citep{correa2018generalized}]\label{th:GA}
Given a causal diagram $\CG$ augmented with selection variable $S$, disjoint sets of variables $\Y,\mathbf{X},\mathbf{Z}$,  for every model compatible with $\mathcal{G}$, we have   
\begin{equation}
\small
     P(\y | do(\mathbf{x})) = \sum_{\Z} P(\y|\mathbf{x},\z,S=1)P(\z |  S=1)
     \label{eq: adjustment}
\end{equation}
if and only if the adjustment variable set $\Z$ satisfies the four criterion shown in  \citep{correa2018generalized}.
\label{thm: GACQ}
\end{theorem}
Instead of identifying causal effect in presence of selection bias by adjustment, \cite{correa2019identification} proposed a parallel procedure to justify whether a causal quantity is identifiable and recoverable from selection bias using axiomatical c-components factorization  \citep{DBLP:conf/aaai/TianP02}. 
However, both techniques require strong graphical condition to obtain the unbiased estimation of the true conditional causal effect when both confounding and selection biases exist.
Basically, c-component factorization first partitions nodes in $\mathcal{G}$ into a set of c-components, then expresses the target intervention in terms of the c-factors corresponding to each c-component. Specifically, a \textit{c-component} $\bm{C}$ denotes a subset of variables in $\mathcal{G}$ such that any two nodes in $\bm{C}$ are connected by a path entirely consisting of bi-directed edges. A bi-directed edge indicates there exists unobserved confounder(s) between the two connected nodes. A \textit{c-factor} $Q[\bm{C}](\mathbf{v})$ is a function that demonstrates the post-intervention distribution of $\bm{C}$ after conducting interventions on the remaining variables $\V \backslash \bm{C}$ and is defined as
\[ Q[\bm{C}](\mathbf{v}) = P(\bm{c}|do(\mathbf{v} \backslash \bm{c})) =  \sum_{\BoldU} \prod_{  V \in \bm{C} }P(v|Pa(v),\mathbf{u}_{v})P(\mathbf{u}) \] 
where $Pa(v)$ and $\mathbf{u}_{v}$ denote the set of observed and unobserved parents for node $V$.
We explicitly denote $Q[\bm{C}](\mathbf{v})$ as $Q[\bm{C}]$ and list the factorization formula.

\begin{theorem}[C-component Factorization \citep{DBLP:conf/aaai/TianP02}]
Given a causal graph $\mathcal{G}$, the target intervention $P(\mathbf{y}|do(\x))$ could be expressed as a product of c-factors associated with the c-components as follows:
\begin{equation}
P(\mathbf{y}|do(\x)) = \sum_{\bm{C} \backslash \Y} Q[\bm{C}] = \sum_{\bm{C} \backslash \Y} \prod_{i = 1}^{l} Q[\bm{C}_i] 
\label{eq: q-decomposition}
\end{equation}    
where $\X, \Y \subset \V$  could be arbitrary sets,  $\bm{C} = An(\Y)_{\CG_{\V \backslash \X}}$ denotes the ancestor node set of $\Y$ in sub-graph $\CG_{\V \backslash \X}$, and $\bm{C}_1, ..., \bm{C}_l$ are the c-components of $\CG_{\bm{C}}$.
\end{theorem}
\cite{bareinboim2014recovering} showed that 
$P(\mathbf{y}|do(\x))$ is recoverable and could be computed by Equation \ref{eq: q-decomposition} if each factor $Q[\bm{C}_i]$ is recoverable from the observational data. Accordingly, they developed the RC algorithm to determine the recoverability of each c-factor.

\section{Related Works}
{\bf \noindent Causal Inference under Confounding and Selection Biases.} 
\cite{bareinboim2014recovering} firstly studied the use of covariate adjustment for simultaneously dealing with both confounding and selection biases based on the SCM. 
\cite{correa2017causal} developed a set of complete conditions to recover causal effects in two cases: when none of the covariates are measured externally, and when all of them are measured without selection bias.  \cite{correa2018generalized} further studied a general case when only a subset of the covariates require external measurement. 
They developed adjustment-based technique that combines the partial unbiased data with the biased data to produce an estimand of the causal effect in the overall population. Different from these works that focus on recovering causal effects under certain graph conditions, our work focuses on bounding causal effects under compound biases, which is needed in various application domains.\\

{\noindent\bf Combining Offline Evaluation and Online Learning in Bandit Setting.}
Recently there are research works that focus on confounding issue in bandit setting \citep{bareinboim2015bandits,tennenholtz2021bandits}. It is shown in \citep{bareinboim2015bandits} that in MAB problems, neglecting unobserved confounders will lead to a sub-optimal arm selection strategy. They also demonstrated that one can not simulate the optimal arm-picking strategy by a single data collection procedure, such as pure offline or online evaluation.   
To this end, another line of research works considers combining offline causal inference techniques and online bandit learning to approximate a nearly-optimal policy.
\cite{tennenholtz2021bandits} studied a linear bandit problem where the agent is provided with partially observed offline data.
\cite{zhang2017transfer, zhang2021bounding} derived causal bounds based on structural causal model and used them to guide arm selection in online bandit algorithms.
\cite{sharma2020warm} further leveraged the information provided by the lower bound of the mean reward to reduce the cumulative regret. Nevertheless, none of the bounds derived by these methods are based on a feature-level causal graph extracted from the offline data.  \cite{li2021unifying, tang2021robust} proposed another direction to unify offline causal inference and online bandit learning by extracting appropriate logged data and feed it to online learning phase. Their VirUCB-based framework mitigates the cold start problem and can thus boost the learning speed for a bandit algorithm without any cost on the regret.
However, none of those proposed algorithms take selection bias and confounding bias simultaneously into consideration during offline evaluation phase.

\section{Algorithm Framework}
\begin{figure}[htpb]
\centering
  \includegraphics[width=0.48\textwidth,height=7.5cm]{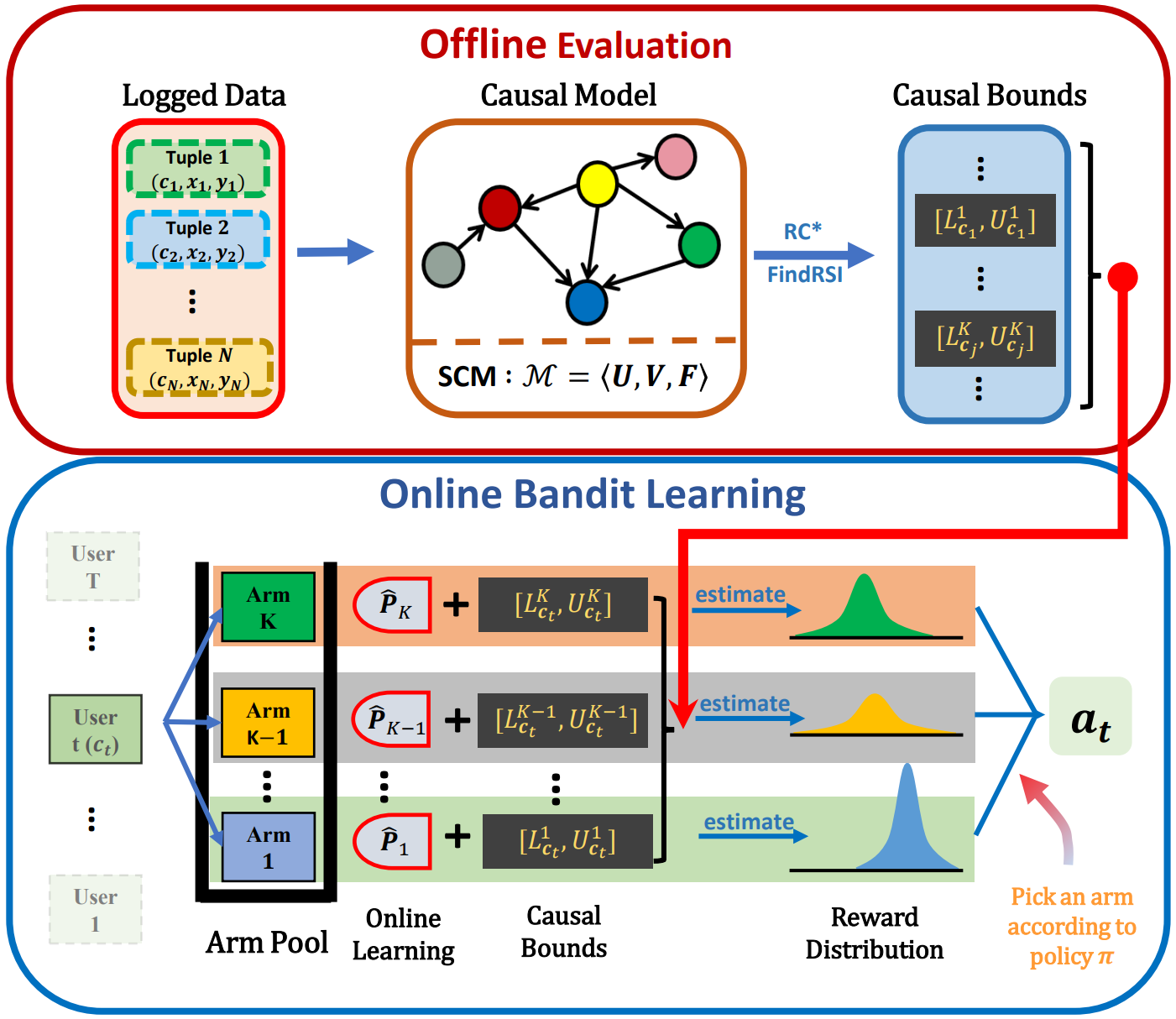}
  \caption{An illustration graph of our proposed framework. }
  \label{fig: illustration}
\end{figure}

An overview of our framework is illustrated in Figure \ref{fig: illustration}. Our algorithm framework leverages the observational data to derive a prior causal bound for each arm to mitigate the cold start issue in online bandit learning, thus reducing the cumulative regret. 
In the offline evaluation phase,  we call our bounding conditional causal effect (BCE) algorithm (shown in Algorithm \ref{alg:conditional causal effect calculation}) to obtain the prior causal bound for each arm given a user's profile. Then in the online bandit phase, we apply adapted contextual bandit algorithms with the prior causal bounds 
as input.

Let $\BC \in \mathcal{C}$ denote the context vector, where $\mathcal{C}$ denotes the domain space of $\BC$.
We use $Y$ to denote the reward variable and $\X \in \mathcal{X}$ to denote the intervention variables. At each time $t \in [T]$, a user arrives and the user profile $\mathbf{c}_t$ is revealed to the agent. The agent pulls an arm $a_t$ with features $\x_{a_t}$ based on previous observations.
The agent then receives the reward $Y_{t}$ and observes values for all non-intervened variables. We define the the expected mean reward of pulling an arm $a$ with  with feature value $\x_{a}$ given user context $\mathbf{c}$ as the conditional causal effect shown below:
\begin{equation}
u_{a,\mathbf{c} } =   \mathbb{E}[Y|do(\X = \x_{a}), \mathbf{c} ]     
\end{equation}



 
When the offline data are available, we can leverage them as prior estimators of the reward for each arm to reduce explorations in the online phase. However, under the circumstances that the causal effect is either unidentifiable or nonrecoverable from the observational data, blindly using the observational data might even have a negative effect on the online learning phase.  Our approach is to derive a causal bound for the desired causal effect from the biased observational data. 
We will further show even when the observational data could only lead to loose causal bounds, we can still guarantee our approach is no worse than conventional bandit algorithms. 

\section{Deriving Causal Bounds under Confounding and Selection Biases}

In this section, we focus on bounding the effects of conditional interventions in the presence of confounding and sample selection biases. To tackle the identifiability issue of a conditional intervention $P(Y=y|do(\x),\mathbf{c})$, the cond-identify algorithm  \citep{DBLP:conf/uai/Tian04} provides a complete procedure to compute conditional causal effects using observational distributions. 
\begin{equation}
P(Y=y|do(\x),\mathbf{c}) = P_{\x}(y|\mathbf{c}) =  \frac{P_{\x}(y,\mathbf{c})}{P_{\x}(\mathbf{c})}  
\label{eq: conditional intervention}
\end{equation}
where $P_{\x}(y|\mathbf{c})$ is the abbreviated form of the conditional causal effect $P(Y=y|do(\x),\mathbf{c})$. \cite{DBLP:conf/uai/Tian04} showed that if the numerator $P_{\x}(y,\mathbf{c})$  is identifiable, then $P_{\x}(y|\mathbf{c})$ is also identifiable. 
In the contextual bandit setting, because none of the variables in $\BC$ is a descendant of variables in $\X$, the denominator $P_{\x}(\mathbf{c})$ can be reduced to $P(\mathbf{c})$ following the causal topology. 
Since $P(\mathbf{c})$ is always identifiable and can be accurately estimated, we do not need to consider the situation 
where neither $P_{\x}(y,\mathbf{c})$ nor $P_{\x}(\mathbf{c})$ is identifiable but $P_{\x}(y|\mathbf{c})$ is still identifiable. Thus the conditional causal effect $P_{\x}(y|\mathbf{c})$ in Equation \ref{eq: conditional intervention} is identifiable \textbf{if and only if} $P_{\x}(y,\mathbf{c})$ is identifiable.  
However, the cond-identify algorithm  \citep{DBLP:conf/uai/Tian04} is not applicable for the scenario with the presence of selection bias.

Algorithm \ref{alg:conditional causal effect calculation}  shows our algorithm framework of bounding conditional causal effects under confounding and selection biases. We develop two methods,  c-component factorization and substitute intervention, and apply each to derive a bound for conditional causal effect separately. We then compare the two causal bounds and return the tighter upper/lower bound. Specifically, lines 5-10 in Algorithm \ref{alg:conditional causal effect calculation} decompose the target causal effect following c-component factorization and recursively call our RC* algorithm (shown in Algorithm \ref{alg:RC}) to bound each c-factor. Lines 11-15  search over recoverable intervention space and find valid substitute interventions to bound the target causal effect. Lines 16-18  compare two derived causal bounds and take the tighter upper/lower bound as the output causal bounds. \textbf{We also include discussions regarding the assumptions on causal graph in Appendix.}

\begin{algorithm}[h]
	\caption{Bounding Conditional Causal Effect}
	\begin{algorithmic}[1]
	\STATE \textbf{function}  BCE$(\x, \mathbf{c} , y,  \mathcal{G}, \mathcal{H})$
	\STATE \textbf{Input:} Intervention variables $\X = \x$, context vector $\BC = \mathbf{c}$, outcome variable $Y = y$, causal graph $\mathcal{G}$.
    \STATE \textbf{Output:} Causal bound $[L_{\x, \mathbf{c}}, U_{\x, \mathbf{c}}]$ of the conditional intervention $P_{\x}(y| \mathbf{c})$.
    \STATE \textbf{Initialization:} $[L_q,U_q] = [0,1], [L_w,U_w] = [0,1]$    
    \smallskip
    \STATE \textcolor{blue}{// C-component Factorization}

    \STATE Decompose $P_{\x}(y, \mathbf{c}) =  \sum_{\D \backslash \{\Y,\BC\} } \prod_{i = 1}^{l} Q[\D_i]$ following Equation \ref{eq: q-decomposition}. 
	\FOR{each $\D_i$}
	    \STATE $L_{Q(\D_i)}, U_{Q(\D_i)} =  \text{RC*}(\D_i,P(\mathbf{v}|S = 1),\CG)$
	\ENDFOR

	\STATE Update $L_q,U_q$ according to Theorem \ref{thm: RCbound}.
    \smallskip
    \STATE \textcolor{blue}{// Substitute Intervention}
    \STATE $\mathcal{D} = \text{FindRSI}(\x, \mathbf{c}, y,\mathcal{G})$
    \IF{$\mathcal{D} \neq \phi$}
    \STATE Update $L_w, U_w$ according to Theorem \ref{thm: sibound}.
    \ENDIF
    \smallskip
	
        \STATE \textcolor{blue}{// Comparing Bounds}
	\STATE Calculate estimated values $\hat{L}_q,\hat{L}_w,\hat{U}_q,\hat{U}_w$ based on $\mathcal{H}$.
	
	\RETURN  $L_{\x, \mathbf{c}} = \max\{\hat{L}_q, \hat{L}_{w} \}$,  $U_{\x, \mathbf{c}} = \min\{\hat{U}_q, \hat{U}_{w} \}$
	\end{algorithmic}
	\label{alg:conditional causal effect calculation}
\end{algorithm}

\subsection{Bounding via C-component Factorization} 
To derive the causal bound based on c-component factorization, we decompose the target intervention into c-factors and call RC* algorithm to recover each c-factor. The RC* algorithm shown in Algorithm \ref{alg:RC}
is designed based on the RC algorithm in \citep{correa2019identification} to accommodate for non-recoverable situations. When the c-factor  $Q[\mathbf{E}]$ is recoverable, the RC* algorithm returns an expression of $Q[\mathbf{E}]$ using biased distribution $P(\mathbf{v}|S=1)$.

Specifically, lines 4-6 in Algorithm \ref{alg:RC} marginalize out the non-ancestors of $\mathbf{E} \cup S$ since they do not affect the recoverability results. From Lemma 3 in \citep{bareinboim2015recovering}, each c-component in line 7 is recoverable since none of them contains ancestors of $S$. Line 13  calls the Identify function proposed by \citep{tian2003identification} that gives a complete procedure to determine the identifiability of $Q[\mathbf{E}]$. When $Q[\mathbf{E}]$ is identifiable, $\text{Identify}(\mathbf{E}, \bm{C}_i, Q[\bm{C}_i])$ returns a closed form expression of $Q[\mathbf{E}]$ in terms of $Q[\bm{C}_i]$.  In line 15, if none of the recoverable c-components $\bm{C}_i$ contains $\mathbf{E}$, we replace the distribution $P$ by dividing the recoverable quantity $\prod_i  Q[\bm{C}_i]$ and recursively run the RC* algorithm on the graph $\mathcal{G}_{\V \backslash \bm{C}}$. Under certain situations where line 8 in RC* Algorithm fails ($\bm{C} = \emptyset$), the corresponding $Q[\mathbf{E}]$ can not be computed from the biased observational data in theory. These situations are referred to as nonrecoverable situations. We address this nonrecoverable challenge by non-parametrically bounding the targeted causal quantity. In this case, RC* returns a bound $[ L_{Q(\mathbf{E})}, U_{Q(\mathbf{E})} ] $ for $Q[\mathbf{E}]$. The bound for $P_{\x}(y, \mathbf{c})$ is derived by summing up the estimator/bounds of those c-factors following Equation \ref{eq: q-decomposition}.


\begin{algorithm}[htpb]
	\caption{RC* Algorithm}
	\begin{algorithmic}[1]
	    \STATE \textbf{function}  RC*$(\mathbf{E}, P , \mathcal{G})$
		\STATE \textbf{Input:} $\mathbf{E}$ a c-component, $P$ a distribution and $\mathcal{G}$ a causal graph.   
		\STATE \textbf{Output:}  Causal bound $[L_{Q[\mathbf{E}]} , U_{Q(\mathbf{E})}]$ for  $Q[\mathbf{E}]$.
		\IF{$\V \backslash (An(\mathbf{E}) \cup An(S)) \neq   \phi$}
		    \RETURN RC*$(\mathbf{E}, \sum_{\V \backslash (An(\mathbf{E}) \cup An(S))}P, \mathcal{G}_{(An(\mathbf{E}) \cup An(S))})$
		\ENDIF 
		\STATE Let $\bm{C}_1, ... , \bm{C}_k$ be the c-components of $\mathcal{G}$ that contains no ancestors of $S$ and $\bm{C} = \cup_{i \in [k]} \bm{C}_i$.
		\IF{$ \bm{C} = \emptyset$}
		    \STATE Bound $Q[\mathbf{E}]$ with $U_{Q(\mathbf{E})} = 1, L_{Q(\mathbf{E})}= 0 $.
                \RETURN $L_{Q(\mathbf{E})},U_{Q(\mathbf{E})}$
		\ENDIF
		\IF {$\mathbf{E}$ is a subset of some $\bm{C}_i$ and $\text{Identify}(\mathbf{E}, \bm{C}_i, Q[\bm{C}_i])$ does not return FAIL}
             \RETURN $L_{Q(\mathbf{E})} = U_{Q(\mathbf{E})} = \text{Identify}(\mathbf{E}, \bm{C}_i, Q[\bm{C}_i])$
        \ENDIF
        \RETURN RC*$(\mathbf{E}, \frac{P}{\prod_i  Q[\bm{C}_i]}, \mathcal{G}_{\V \backslash \bm{C}})$
	
	\end{algorithmic}
	\label{alg:RC}
\end{algorithm}

Note that in line 9 of RC* algorithm, we bound the target c-component by $[0,1]$ since under semi-Markovian models it is challenging to find a tight bound for $Q[\mathbf{E}]$ when $\bm{C} = \emptyset$. One future direction is to further apply a non-parametric bounding technique similar to \citep{wu2019counterfactual}. That is, choosing certain probability distributions in the truncated formula that are the source of unrecoverability, and set specific domain values for a carefully chosen variable set to allow these distributions achieve their maximum/minimum values. Finally we list the causal bounds derived from calling RC* algorithm in the follow Theorem:
\begin{theorem}[Causal Bound from RC* algorithm]
    Given a conditional intervention $P_{\x}(y| \mathbf{c})$, the causal bounds derived by calling RC* algorithm for each c-factor are:
\begin{equation}
    \begin{split}
    L_q = \sum_{\D \backslash \{\Y,\BC\} } \prod_{i = 1}^{l} L_{Q[D_i]} / P_{\x}(\mathbf{c}) \\
    U_q =  \sum_{\D \backslash \{\Y,\BC\} } \prod_{i = 1}^{l} U_{Q[D_i]} / P_{\x}(\mathbf{c})  
    \end{split}
\label{eq: RC*}
\end{equation}
\label{thm: RCbound}
\end{theorem}

\subsection{Bounding via Substitute Interventions}
From previous discussion we find that RC* algorithm may return a loose bound when we fail to recover most of the c-factors. In order to obtain a tight causal bound that is robust under various graph conditions, we develop another novel strategy to bound $P_{\x}(y, \mathbf{c})$. Our key idea is to search over the substitute recoverable interventions with a larger intervention space. Note that for a variable set $\W$ such that $\W \cap \X = \emptyset$
in the contextual bandit setting, we can perform marginalization on $\W$ and derive
$P_{\x}(y, \mathbf{c}) = \sum_{\mathbf{\W}}P_{\x}(y, \mathbf{c}|\w)P(\w)$.
We can further bound $P_{\x}(y, \mathbf{c})$  by
\begin{equation}
\begin{split}
    P_{\x}(y, \mathbf{c}) \leq \max\limits_{\mathbf{w}^* \in \W }P_{\x}(y, \mathbf{c}|\w^*)\\
   P_{\x}(y, \mathbf{c}) \geq \min\limits_{\mathbf{w}^* \in \W }P_{\x}(y, \mathbf{c}|\w^*)
\end{split}
\end{equation}
We then investigate whether the action/observation exchange rule of do-calculus  \citep{pearl2009causality} and the corresponding graph conditions could be extended in the presence of selection bias and list the results in the following lemma.
\begin{lemma}[Action/Observation Rule under Selection Bias]
If the graphical condition $(\Y \independent \Z, S | \X, \W)_{\mathcal{G}_{ \overline {\X} \overline{\Z(\w)}}}$ is satisfied in $\mathcal{G}$, the following equivalence between two post-interventional distributions holds:
\begin{equation}
        P(\y|do(\x),do(\w),\z, S=1) = P(y|do(\x),\W,\z,S=1)
\label{eq: Action/Observation Rule}
\end{equation}
where $\mathcal{G}_{\overline {\X}\underline{\Z}}$  represents the causal graph with the deletion of both incoming and outgoing arrows of $\X$ and $\Z$ respectively.
$\Z(\W)$ is the set of $\Z$-nodes that are not ancestors of variables in $\W$ in $\mathcal{G}_{\overline {\X}}$.
\label{lemma: rule 2}
\end{lemma}

Following the general action/observation exchange rules in Equation \ref{eq: Action/Observation Rule}, if  $(Y, \BC \independent \W, S | \X)_{\mathcal{G}_{\overline {\X}\underline{\W}}}$, we can replace $P_{\x}(y, \mathbf{c}|\w^*)$ with $P_{\x,\w^*}(y, \mathbf{c})$ and derive the bound for $P_{\x} (y,\mathbf{c})$ as shown in Theorem \ref{thm: sibound}.
\begin{theorem}[Causal Bound with Substitute Interventions]
Given a set of variables corresponding to recoverable substitute interventions:  $\mathcal{D} = \{  \W  |   P_{\mathbf{x}, \w}(y,\mathbf{c}) ~ \text{is recoverable}\}$, the target conditional intervention $P_{\x}(y| \mathbf{c})$ is bounded by 
\begin{equation}
    \begin{split}
    L_{w} = \max\limits_{\W \in \mathcal{D}}\min\limits_{ \w^* \in \W} P_{\mathbf{x},\w^*}(y,\mathbf{c}) / P_{\x}(\mathbf{c}) \\
    U_{w} = \min\limits_{\W \in \mathcal{D}}\max\limits_{ \w^* \in \W} P_{\mathbf{x},\w^*}(y,\mathbf{c}) / P_{\x}(\mathbf{c}) 
    \end{split}
\label{eq: si}
\end{equation}
\label{thm: sibound}
\end{theorem}

\begin{algorithm}[ht]
	\caption{Finding Recoverable Substitute Interventions}
	\begin{algorithmic}[1]
	\STATE \textbf{function}  FindRSI$(\x, \mathbf{c}, y,\mathcal{G})$
	\STATE \textbf{Input:}  Causal graph $\mathcal{G}$, target intervention $P_{\x}(y,\mathbf{c})$.
	\STATE \textbf{Output:} A valid variable set $\mathcal{D} = \{  \W  |   P_{\mathbf{x}, \w}(y,\mathbf{c})$ is recoverable and could be expressed in terms of biased observational distributions following Equation \ref{eq: adjustment}$\}$.
	\STATE \textbf{Initialize:}  $\mathcal{D} \leftarrow \emptyset$.

	    \FORALL{ $\W$ such that $\W \cap \X = \emptyset$, starting with the smallest size of $\W$}
		    \IF{a valid adjustment set can be found according to Theorem \ref{th:GA}}
    		    \STATE $\mathcal{D} = \mathcal{D} \cup \{ \W \}$
		     \ENDIF
		\ENDFOR
	\end{algorithmic}
	\label{alg:findw}
\end{algorithm}
We list our procedure of finding all the recoverable substitute interventions in Algorithm \ref{alg:findw}. Basically the main function FindRSI in Algorithm \ref{alg:findw} returns a set containing all admissible variables, each of which corresponding to a recoverable intervention with a larger intervention space.

\begin{figure}[t]
	\centering
	\begin{tikzpicture}[scale=1.1]
	\definecolor{blue}{rgb}{0.0, 0.0, 0.0}
	\definecolor{red}{rgb}{0.8, 0.0, 0.0}
	\definecolor{green}{rgb}{0.0, 0.5, 0.0}
	
	\tikzstyle{vertex} = [very thick, circle, draw, inner sep=0pt, minimum size = 8mm]
    \tikzstyle{vertexd} = [very thick, circle, draw, inner sep=0pt, dashed, minimum size = 8mm]
	\tikzstyle{empty} = [very thick, circle, inner sep=0pt, minimum size = 8mm]
	\tikzstyle{rect} = [very thick, rectangle, draw, inner sep=0pt, minimum size = 6mm]
	\tikzstyle{remp} = [very thick, rectangle, inner sep=0pt, minimum size = 3mm]
	\tikzstyle{edge} = [ultra thick, -stealth, blue]
 
	\node[rect] at (3.6 ,3.5) (S) {$S$};
	
	\node[vertex] at (1.3,3.2) (X1) {$X_1$};
	\node[vertex] at (1.3,1.6) (X2) {$X_2$};
	\node[vertex] at (0.0,2.7) (U1) {$U_1$};
	\node[vertex] at (0.0,1.6) (U2) {$U_2$};
	\node[vertex] at (2.6,3.2) (I1) {$I_1$};
    \node[vertexd] at (4.2,2.6) (C1) {$C_1$};
	\node[vertex] at (2.8,2.0) (Y) {$Y$};

	\draw[edge, black] (U1) to (Y);
	\draw[edge, black] (U1) to (X1);
        \draw[edge, black] (U2) to (X2);
        \draw[edge, black] (X2) to (Y);
        \draw[edge, black] (X1) to (I1);
        \draw[edge, black] (I1) to (S);   
        \draw[edge, black] (I1) to (Y);  
        \draw[edge, black] (C1) to (I1);
        \draw[edge, black] (C1) to (Y);
    
	\end{tikzpicture}
	\caption{Causal graph for synthetic data.}
	\label{fig: synthetic}
\end{figure}
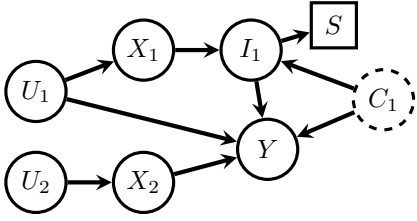

Next, we give an illustration example on how to run our BCE algorithm to get prior causal bounds. Figure \ref{fig: synthetic} shows a causal graph constructed from offline data, where nodes $U_1,U_2$ and $X_1,X_2$ depict user features and item features respectively,  $Y$ denotes the outcome variable, $S$ denotes the selection variable, and $I_1$ denotes an intermediate variable. The dashed node $C_1$ denotes the confounder that affects both $I_1$ and $Y$ simultaneously. 
To bound the conditional causal effect  $p_{x_1,x_2}(y|u_1,u_2)$ via c-component factorization, we first identify the set $\D = An(\Y)_{\CG_{\V \backslash \X}} = \{ Y, U_1,U_2 \}$. The target intervention could be expressed as 
\begin{equation}
    \begin{split}
       p_{x_1,x_2}&(y|u_1,u_2) =  p_{x_1,x_2}(y,u_1,u_2)/p(u_1,u_2) \\
       &\quad\;\;= (Q[Y] \cdot Q[U_1] \cdot Q[U_2])/p(u_1,u_2)  
    \end{split}
\end{equation}
We then call RC* algorithm to bound each c-component and return the bound for each arm according to Theorem \ref{thm: RCbound}. For bounding causal effects via substitute interventions, we call FindRSI to find a valid variable set $\mathcal{D} = \{I_1\}$. According to Theorem \ref{thm: sibound}, we can obtain the bound for each arm. 

\section{Online Bandit Learning with Prior Causal Bounds}

In this section we show how to incorporate our derived causal bounds to online contextual bandit algorithms. We focus on the stochastic contextual bandit setting with linear reward function. 
We define the concatenation of user and arm feature as $\x_{\mathbf{c}_t,a} = [\mathbf{c}_t,\x_a] \in \mathbb{R}^d$ when the agent picks arm $a$ based on user profile $\mathbf{c}_t$ at time $t \in [T]$ and use its simplified notion $\x_{t,a}$ to be consistent with previous work like LinUCB.
Under the linear assumption, the binary reward is generated by a Bernoulli distribution $Y_a \sim Ber(\langle \bm{\theta},\x_{t,a} \rangle)$ parameterized by $\bm{\theta}$.
Let  $a_t = \argmax_{a \in \mathcal{A}} \mathbb{E}[Y_{a}] $, the expected cumulative regret up to time $T$ is defined as: 
\[  \mathbb{E}[ R(T)]=  \sum_{t=1}^{T} \langle \bm{\theta}, \x_{t,a^*} \rangle -  \sum_{t=1}^{T} \mathbb{E}[ Y_{a_t}] \]
At each round the agent pulls an arm based on the user context, observes the reward, and aims to minimize the expected cumulative regret $\mathbb{E}[R(T)]$ after $T$ rounds. We next conduct regret analysis and show our strategy could consistently reduce the long-term regret with the guide of
a prior causal bound for each arm's reward distribution. 

\subsection{LinUCB algorithm with Prior Causal Bounds}

LinUCB  \citep{chu2011contextual} is one of the most widely used stochastic contextual bandit algorithms that assume the expected reward of each arm $a$ is linearly dependant on its feature vector $\mathbf{x}_{t,a}$ with an unknown coefficient  $\bm{\theta}_a$ at time $t$. We develop the LinUCB-PCB algorithm  that includes a modified arm-picking strategy,  clipping the original upper confidence bounds with the prior causal bounds obtained from the offline evaluation phase. Algorithm \ref{alg:linucb-PCB} shows the pseudo-code of our LinUCB-PCB algorithm. The truncated upper confidence bound shown in line 10 of Algorithm \ref{alg:linucb-PCB} contains strong prior information about the true reward distribution implied by the prior causal bound, thus leading to a lower asymptotic regret bound.  \textbf{We include the  proof details of Theorem \ref{thm: linucb_pcb regret} and \ref{thm: reduced regret} in Appendix}.

\begin{algorithm}[h]
	\caption{LinUCB algorithm with Prior Causal Bounds (LinUCB-PCB)}
	\begin{algorithmic}[1]
		\STATE \text{\textbf{Input:}} Time horizon $T$, arm set $\mathcal{A}$, prior causal bounds $\{[L_{a,\mathbf{c}},  U_{a,\mathbf{c}}]\}_{a, \mathbf{c} \in  \{ \mathcal{A}, \mathcal{C}\}}$, $\alpha$ $\in$ $\mathbb{R}^+$.
	    \FOR {t = 1,2,3,...,$T$} 
\STATE Observe $\mathbf{x}_{t,a} \in \mathbb{R}^d$ for user profile $c_t$ and every arm $a$ 
      
	    \FOR {$a \in \mathcal{A}$ }
	    \IF  {$a$ is new}
            \STATE $A_a \leftarrow \mathbf{I_d}$, $\mathbf{b}_a \leftarrow \mathbf{0}_{d \times 1}$
    	\ENDIF
	    \smallskip
	    \STATE $\bm{\hat{\theta}}_a \leftarrow A^{-1}_a \mathbf{b}_a$
	    \STATE $UCB_a(t) \leftarrow \hat{\bm{\theta}}_a^\mathrm{T}\mathbf{x}_{t,a} + \alpha \sqrt{\mathbf{x}_{t,a}^\mathrm{T} A_a^{-1} \mathbf{x}_{t,a}}$
            \STATE $\overline{UCB}_{a}(t) \leftarrow \min \big\{ UCB_a(t),  U_{a,\mathbf{c}_t}\}$
            \STATE  \textcolor{blue}{//Truncated UCB}
	    \ENDFOR
            \STATE Pull arm $a_t  \leftarrow  argmax_{a \in \mathcal{A}} \overline{UCB}_{a}(t)$, and observe a reward  $r_{t,a_t}$
     	\STATE $A_{a_t} \leftarrow A_{a_t} +\mathbf{x}_{t,a_t}\mathbf{x}^\mathrm{T}_{t,a_t}$, $\mathbf{b}_{a_t} \leftarrow \mathbf{b}_{a_t} + r_{t,a_t}\mathbf{x}_{t,a_t} $
	    \ENDFOR
	\end{algorithmic}
	\label{alg:linucb-PCB}
\end{algorithm}


\begin{theorem}
Let $||\mathbf{x}||_2$  define the L-2 norm of a context vector $\mathbf{x} \in \mathbb{R}^d$ and
\[L = \max_{ a,\mathbf{c} \in \{ \mathcal{A}, \mathcal{C}\}, U_{a,\mathbf{c}}\geq\langle \bm{\theta}, \x_{a^*,\mathbf{c}} \rangle}  ||\mathbf{x}_{a,\mathbf{c}}||_2\]
The expected regret of LinUCB-PCB algorithm is bounded by:
\begin{gather*}
\begin{split}
    R_T &\leq  \sqrt{2Tdlog(1+TL^2/(d\lambda))} \\
    &\times 
    2(\sqrt{\lambda}M + \sqrt{2log(1/\delta) + dlog(1+T L^2/(d\lambda))})
\end{split}
\end{gather*}
where $M$ denotes the upper bound of $||\bm{\theta}_a||_2$ for all arms, $\lambda$ denotes the penalty factor corresponding to the ridge regression estimator $\bm{\hat{\theta}}_a$.
\label{thm: linucb_pcb regret}
\end{theorem}

We follow the standard procedure of deriving the expected regret bound for linear contextual bandit algorithms in \citep{abbasi2011improved} and \citep{lattimore2020bandit}. We next discuss the potential improvement in regret that can be achieved by applying LinUCB-PCB algorithm in comparison to original LinUCB algorithm. 

\begin{theorem}
If there exists an arm $a$ such that $U_{a,\mathbf{c}_t} < \langle \bm{\theta}, \x_{a^*,\mathbf{c}_t} \rangle$ at a round $t \in [T]$,  
LinUCB-PCB is guaranteed to achieve lower cumulative regret than LinUCB algorithm.
\label{thm: reduced regret}
\end{theorem}
We further define the total number of sub-optimal arms implied by prior causal bounds as  
\[N_{pcb}^{-} = \sum_{a, \mathbf{c} \in  \{ \mathcal{A}, \mathcal{C}\}}  \mathds{1}_{U_{a,\mathbf{c}} - \langle \bm{\theta}, \x_{a^*,\mathbf{c}} \rangle < 0}\]
Note that the value of $N_{pcb}^{-}$ depends on the accuracy of the causal upper bound for each arm. This is because if the estimated causal bounds are more concentrated, that is, $U_{a,
\mathbf{c}}$ is close to $\langle \bm{\theta}, \x_{a,\mathbf{c}} \rangle$ for each $a, \mathbf{c} \in  \{ \mathcal{A}, \mathcal{C}\}$, there will be more arms whose prior causal upper bound is less than the optimal mean reward, thus $N_{pcb}^{-}$ will increase accordingly. A large $N_{pcb}^{-}$ value implies less uncertainty regarding the sub-optimal arms implied by prior causal bounds. As a result there are in general less arms to be explored and the $L$ value will decrease accordingly, leading to a more significant improvement by applying LinUCB-PCB algorithm.\\

\noindent \textbf{Extension} We have also investigated leveraging the developed causal bounds to further improve one state-of-the-art contextual bandit algorithm \cite{hao2020adaptive} and one classical non-contextual bandit algorithm \cite{lattimore2020bandit}. Due to page limit we defer our developed OAM-PCB and UCB-PCB algorithms as well as the corresponding regret analysis to Appendix.

\section{Empirical Evaluation}
In this section, we conduct experiments to validate our proposed methods. We use the synthetic data generated following the graph structure in Figure \ref{fig: synthetic}. We generate 30000 data points following the conditional probability table in Appendix to simulate the confounded and selection biased setting. After conducting the preferential exclusion indicated by the selection mechanism, there are approximately 15000 data points used for offline evaluation.\\

\begin{figure}[htb!]
\centering
  \includegraphics[width=7cm,height=5cm]{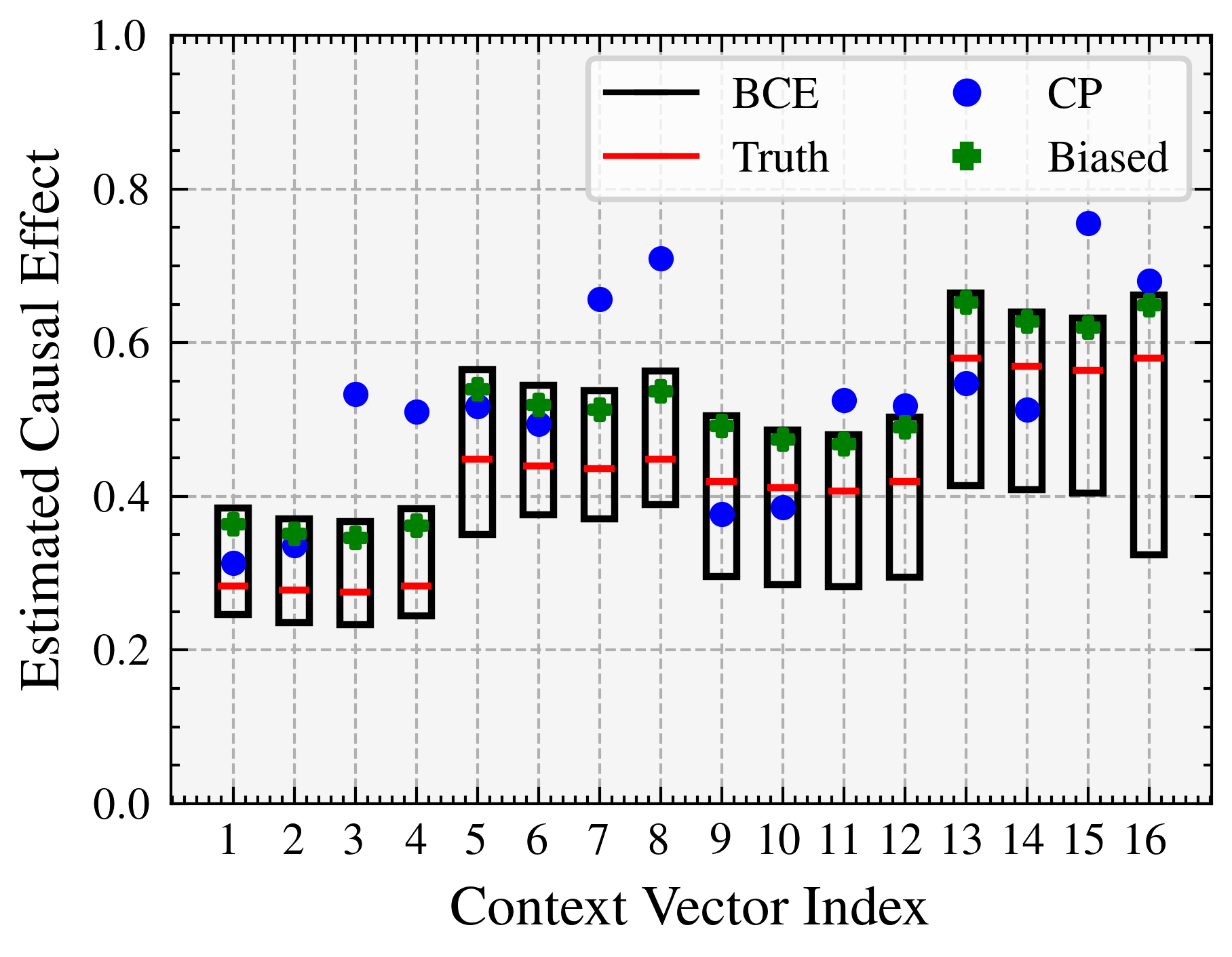}
  \caption{Comparison results for offline evaluation under confounding and selection biases. }
  \label{fig: reward estimation}
\end{figure}

\noindent{\textbf{Offline Evaluation}} 
We use our BCE algorithm to obtain the bound of each arm based on the input offline data and compare the causal bound derived by the algorithm with the estimated values from two baselines: an estimate that is derived based on Equation \ref{eq: backdoor adjustment} which only takes into account confounding bias (Biased), and a naive conditional probability estimate derived without considering both confounding and selection biases (CP).
We further report the causal bounds and the estimated reward among 16 different values of the context vector in Figure \ref{fig: reward estimation}. The comparison results show our BCE algorithm contains the ground-truth causal effect (denoted by the red lines in the figure) for each value of the context vector. On the contrary, the estimated values from CP and Biased baselines  deviate from the true causal effect in the presence of compound biases. The experimental results reveal the fact that neglecting any bias will inevitably lead to an inaccurate estimation of the target causal effect.  

 \begin{figure}[th!]
    \centering
        \centering
        \includegraphics[width=7cm,height=5cm]{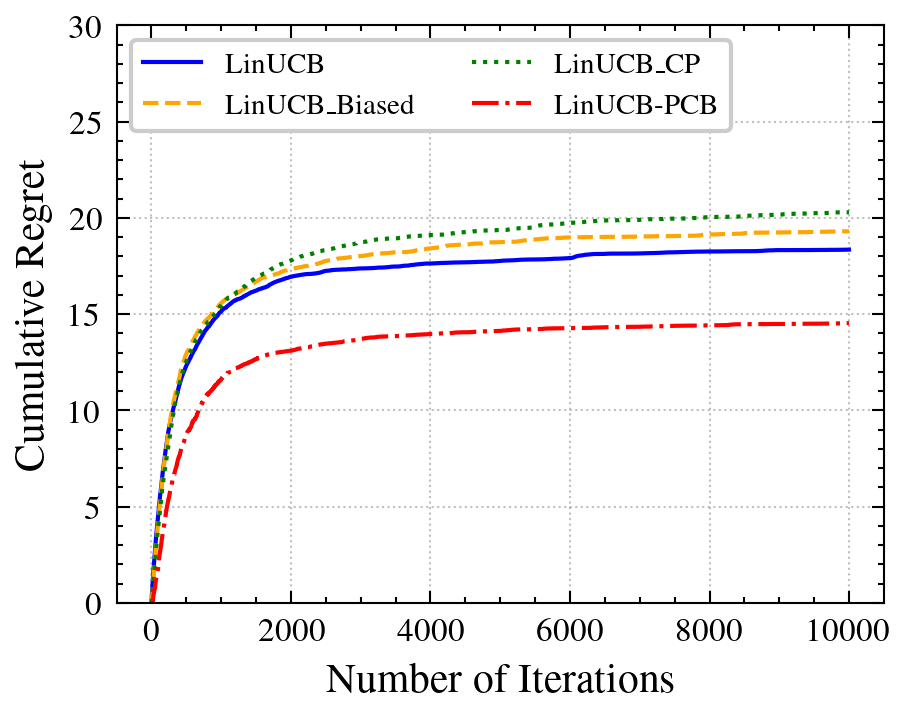}
    \caption{Comparison results for contextual linear bandit.}   
\label{fig: comparison result}
\end{figure}
\noindent{\textbf{Online Bandit Learning}} 
We use 15000 samples from the generated data to simulate the online bandit learning process. In Figure \ref{fig: comparison result}, we compare the performance of our LinUCB-PCB algorithm regarding cumulative regret with the following baselines: LinUCB, LinUCB\textunderscore Biased and LinUCB\textunderscore CP, where LinUCB\textunderscore Biased and LinUCB\textunderscore CP are LinUCB-based algorithms initialized with the estimated reward for each value of the context vector (arm) from the Biased and CP baselines in the offline evaluation phase. Each curve denotes the regret averaged over 100 simulations to approximate the true expected regret. We find that LinUCB-PCB achieves the lowest regret compared to the baselines. Moreover, both LinUCB\textunderscore Biased and LinUCB\textunderscore CP perform worse than the LinUCB baseline, which is consistent with the conclusion from our theoretical analysis that blindly utilizing biased estimates from offline data could negatively impact the performance of online bandit algorithms.  

\section{Conclusion}
This work studies bounding conditional causal effects in the presence of confounding and sample selection biases using causal inference techniques and utilizes the derived bounds to robustly improve online bandit algorithms. We present two novel techniques to derive causal bounds for conditional causal effects given offline data with compound biases. We  develop contextual and non-contextual bandit algorithms that leverage the derived causal bounds and conduct their regret analysis. Theoretical analysis and empirical evaluation demonstrate the improved regrets of our algorithms.
In future work, we will study incorporating causal bounds into advanced bandit algorithms such as state-of-the-art linear contextual bandits \cite{hao2020adaptive}, contextual bandits under non-linearity assumption \cite{zhou2020neural}, and bandits with adversarial feedback \cite{luo2023improved}, to comprehensively demonstrate the generalization ability of our approach.

\clearpage

\section*{Acknowledgements}
This work was supported in part by NSF under grants 1910284, 1940093, 1946391, 2137335, and 2147375.

\section*{Ethics Statement}
Our research could benefit online recommendation system providers so that
they can mitigate biases from multiple sources while conducting recommendations. Our research could also benefit users of online recommendation systems
as we aim to eliminate the influence of biases and achieve accurate personalized recommendation. Since fairness could be regarded as a certain type of bias,
our research could be further extended to prevent users from receiving biased
recommendations, especially for those from disadvantaged groups. To the best
knowledge, we do not see any negative ethical impact from our paper.

\bibliography{aaai24}

\clearpage

\appendix

\section{Assumptions on Causal Graph, Notations and Proofs}
In this paper we develop novel approaches for deriving bounds of conditional causal effects in the presence of both confounding and selection biases. Our approach allows the existence of unobserved confounders, which are denoted using dashed bi-directed arrows in the causal graph $\mathcal{G}$.  We also introduce the selection node $S$ depicting the data selection mechanism in the offline evaluation phase.  With slight abuse of the notation, after the background section we use $\mathcal{G}$ to denote the causal graph augmented with $S$ for simplicity. 
Following the common sense in this research subfield \cite{lu2020regret,correa2017causal,zhang2017transfer,lee2018structural} we assume the causal graph is accessible by the agent by adopting state-of-the-art causal discovery techniques  and remains invariant through the offline evaluation phase and online learning phase. 

Throughout the proofs, we use bold letters to denote a vector.  We use $||\mathbf{x}||_2$ to define the L-2 norm of a vector $\mathbf{x} \in \mathbb{R}^d$. For a positive definite matrix $A \in \mathbb{R}^{d \times d}$, we define the weighted 2-norm of $\mathbf{x} \in \mathbb{R}^d$ to be $||\mathbf{x}||_A = \sqrt{\mathbf{x}^\mathrm{T}A\mathbf{x}}$.

\section{Proof of Theorem 5}
To derive the regret bound of LinUCB-PCB algorithm, we follow existing research works (e.g., \citep{abbasi2011improved,wu2016contextual}) to make four common assumptions defined as follows:
\begin{enumerate}
    \item The error term $\epsilon_t$ follows 1-sub-Gaussian distribution for each time point. 
    \item$\{\alpha_t\}^n_{i=1} $ is a non-decreasing sequence with $\alpha_1 \geq 1$.
    \item $||\bm{\theta}^*||_2 < M $ 
    \item There exists a $\delta \in (0,1)$ such that with probability $1-\delta$, for all $t \in [T], \bm{\theta}^* \in \mathcal{C}_t$ where $\mathcal{C}_t$ satisfies Equation \ref{simplified_bound}.
\end{enumerate}

We begin the proof by introducing four technical lemmas from \citep{abbasi2011improved} and \citep{lattimore2018bandit} as follows:

\begin{lemma}
(Theorem 2 in \citep{abbasi2011improved}) Suppose the noise term is 1-sub-Gaussian distributed, let $\delta \in (0,1)$, with probability at least $1-\delta$, it holds that for all $t \in \mathbb{N^+}$,
\begin{equation}
\begin{split}
 ||\bm{\theta}^* - \bm{\hat{\theta}}_t ||_{A_t} \leq \sqrt{\lambda}||\bm{\theta}^*||_2 +  \sqrt{2log(|A_t|^{1/2}|\lambda I_d|^{-1/2}\delta^{-1}) }
\end{split}
\label{confidence_bound}
\end{equation}
\label{abbasi}
\end{lemma}
\begin{lemma}
(Lemma 11 in appendix of \citep{abbasi2011improved}) If $\lambda \geq max(1,L^2)$, the weighted L2-norm of feature vector could be bounded by :
\[\sum_{t=1}^{T}||\mathbf{x}_{t,a}||^2_{A^{-1}_t}  \leq 2 log \frac{|A_t|}{\lambda^d}\]
\label{zhou_lemma1}
\end{lemma} 

\begin{lemma}
(Lemma 10 in appendix of \citep{abbasi2011improved} ) The determinant $|A_t|$ could be bounded by:
$|A_t| \leq (\lambda + t L^2/d)^d$.
\label{zhou_lemma2}
\end{lemma}

\begin{lemma}
 (Theorem 20.5 in \citep{lattimore2018bandit}) With probability at least $1-\delta$, for all the time point $t \in \mathbb{N}^+$ the true coefficient $
\bm{\theta}^*$ lies in the set:
\begin{equation}
\begin{split}
\mathcal{C}_t  = \{ &\bm{\theta} \in \mathbb{R}^d :||\bm{\hat{\theta}}_t - \bm{\theta}||_{A_t} \leq \\
&\sqrt{\lambda}M + \sqrt{2log(1/\delta) + dlog(1+T L^2/(d\lambda))} \} 
\end{split}
\label{simplified_bound}
\end{equation}
\label{alpha_T}
\end{lemma}

According to the arm selection strategy and OFU principle, the regret at each time $t$ is bounded by: 
\begin{gather*}
\begin{split}
reg_t  
& = \mathbf{x}_{t,a}^\mathrm{T}\bm{\hat{\theta}}_t -
\mathbf{x}^\mathrm{T}_{t,a}\bm{\theta}^* \\
&\leq \mathbf{x}_{t,a}^\mathrm{T}\bm{\hat{\theta}}_t + \alpha_t||\mathbf{x}_{t,a}||_{A^{-1}_t}  - \mathbf{x}^\mathrm{T}_{t,a}\bm{\theta}^*\\
&\leq \mathbf{x}_{t,a}^\mathrm{T}\bm{\hat{\theta}}_t + \alpha_t||\mathbf{x}_{t,a}||_{A^{-1}_t}
- (\mathbf{x}^\mathrm{T}_{t,a}\bm{\hat{\theta}_t} -\alpha_t||\mathbf{x}_{t,a}||_{A^{-1}_t})   \\
&\leq 2\alpha_t||\mathbf{x}_{t,a}||_{A^{-1}_t}\\
\end{split}
\label{eq:r_t}
\end{gather*}
Summing up the regret at each bound, with probability at least $1-\delta$ the cumulative regret up to time $T$ is bounded by:
\begin{equation}
\begin{split}
R_T &= \sum_{t=1}^{T}reg_t \leq \sqrt{T\sum_{t=1}^{T}reg_t^2}
\leq 2\alpha_T\sqrt{T\sum_{t=1}^{T}||\mathbf{x}_{t,a}||^2_{A^{-1}_t}}  
\end{split}
\label{eq:R_t}
\end{equation}

Since $\{\alpha_t\}^n_{i=1}$ is a non-decreasing sequence, we can enlarge each element $\alpha_t$ to $\alpha_T$ to obtain the inequalities in Equation \ref{eq:R_t}.
By applying the inequalities from Lemma \ref{zhou_lemma1} and \ref{zhou_lemma2} we could further relax the regret bound up to time $T$ to:
\begin{equation}
\begin{split}
R_T &\leq 2\alpha_T\sqrt{2Tlog\frac{|A_t|}{\lambda^d}}\\
&\leq 2\alpha_T\sqrt{2Td(log(\lambda+TL^2/d)-log\lambda)}\\
&= 2\alpha_T\sqrt{2Tdlog(1+TL^2/(d\lambda))}
\end{split}
\label{eq:R_t_further}
\end{equation}

Following the result of Lemma \ref{abbasi}, by loosing the determinant of $A_t$ according to Lemma \ref{zhou_lemma2},
Lemma \ref{alpha_T} provides a suitable choice for $\alpha_T$ up to time $T$. By plugging in the RHS from Equation \ref{simplified_bound} we derive the cumulative regret bound:
\begin{gather*}
\begin{split}
    R_T &\leq  \sqrt{2Tdlog(1+TL^2/(d\lambda))} \\
    &\times 
    2(\sqrt{\lambda}M + \sqrt{2log(1/\delta) + dlog(1+T L^2/(d\lambda))})
\end{split}
\end{gather*}

\section{Proof of Theorem 6}
    To prove Theorem 6, we first introduce a Lemma shown as follows:
\begin{lemma}[Reduced Arm Exploration Set]
    Given an arm $a$ with $U_{a,\mathbf{c}} <  \langle \bm{\theta}, \x_{a^*,\mathbf{c}} \rangle$, we have $P(a_t = a) = 0, \forall t \in [T]$.
\label{lemma: reduced set}
\end{lemma}
We prove Lemma \ref{lemma: reduced set} by contradiction. Given an arm $a$ with $U_{a,\mathbf{c}} <  \langle \bm{\theta}, \x_{a^*,\mathbf{c}} \rangle$ and suppose the agent pulls arm $a$ at round $t$. Based on the definition of $\overline{UCB}_{a,\mathbf{c}} = \min \big\{ UCB_{a}(t),  U_{a,\mathbf{c}}\}$ and the optimism in the face of uncertainty (OFU) principle we have $\langle \bm{\theta}, \x_{a^*,\mathbf{c}} \rangle < \overline{UCB}_{a^*,\mathbf{c}} < \overline{UCB}_{a,\mathbf{c}} \leq U_{a,\mathbf{c}}$, which contradicts with the fact $U_{a,\mathbf{c}} <  \langle \bm{\theta}, \x_{a^*,\mathbf{c}} \rangle$. Thus  $\forall t \in [T]$ we have $P(a_t = a) = 0$. Lemma \ref{lemma: reduced set} basically states that although the optimal reward given a user context is unknown to the agent apriori, based on the exploration strategy enhanced with the information provided by the prior causal bounds, LinUCB-PCB will not pull the sub-optimal arms implied by their upper causal bounds at each round, thus leading to a reduced exploration arm set and a lower value of $L$.

\section{Implementation Details of OAM-PCB Algorithm}
\cite{hao2020adaptive} recently developed  one state-of-the-art contextual linear bandit algorithm based on the optimal allocation matching (OAM) policy. It alternates between exploration and exploitation based on whether or not all the arms have satisfied the approximated allocation rule. We investigate how to incorporate prior causal bounds in OAM and develop the new OAM-PCB algorithm. 

As shown in Algorithm \ref{alg: oam}, at each round in both exploitation and wasted exploration scenarios, we truncate the upper confidence bound for each arm with the upper causal bound to obtain  a more accurate estimated upper bound:

\begin{equation}
\begin{split}
\widehat{UCB}_a(t-1) &= \min \Bigl\{ a^\top \hat{\theta}_{t-1} + \sqrt{f_{n,1/s(t))^2}} ||a||_{G^{-1}_{t-1}}, U_{a,\mathbf{c}}  \Bigl\}\\
a_t &= \argmax_{a \in \mathcal{A}}\widehat{UCB}_a(t-1)
\end{split}
\label{eq: estimated ucb}
\end{equation}
where $G_t = \sum_{s=1}^t \X_s \X_s^\top$.
The algorithm then explores by computing two arms:
\begin{equation}
\begin{split}
&b_1 = \argmin_{a \in \mathcal{A}} \frac{N_a(t-1)}{\min( T_a^{c_t}(\widehat{\Delta}(t-1)), f_n/\widehat{\Delta}^2_{min}(t-1))}   \\
&b_2 = \argmin_{a \in \mathcal{A}} N_a(t-1)
\end{split}
\label{eq: arm exploration}
\end{equation}
where $f_{n,\delta} = 2(1 + 1/log(n))log(1/\delta) + cdlog(dlog(n))$. $c$ is a constant and we denote $f_n = f_{n,1/n}$ for simplicity.  $N_a(T)$ denotes the number of pulls of arm $a$ up to time $T$. For any $\Tilde{\Delta} \in [0,\infty)^{|\cup_{a \in \mathcal{A}}|}$ that is an estimate of $\Delta$,  $T(\Tilde{\Delta})$ could be treated as an approximated allocation rule in contrast to the optimal allocation rule, which is defined as a solution to the following optimization problem:
\begin{equation}
    \min_{(T^c_a)_{a,c} \in[0,\infty]} \sum_{c=1}^{|\mathcal{C}|}\sum_{a \in \mathcal{A}}T^c_a \Tilde{\Delta}^c_a
\end{equation}
subject to
\begin{equation}
    ||x||^2_{H_T^{-1}} \leq \frac{\Delta_{a}^2}{f_n}, \forall a \in \mathcal{A}, c \in [|\mathcal{C}|]
\end{equation}
and $H_T = \sum_{c=1}^{|\mathcal{C}|}\sum_{a \in \mathcal{A}}T^c_a a a^\top$ is invertible.\\

\begin{algorithm}[h]
	\caption{Optimal Allocation Matching with Prior Causal Bounds}
	\begin{algorithmic}[1]
		\STATE \textbf{Input:} Time horizon $T$, arm set $\mathcal{A}$, exploration parameter $\epsilon_t$, exploration counter $s(d) = 0$, prior causal bounds $\{[L_{a,\mathbf{c}},  U_{a,\mathbf{c}}]\}_{a, \mathbf{c} \in  \{ \mathcal{A}, \mathcal{C}\}}$.
           \FOR{$t = 1 ~ \textbf{to} ~ T$} 
           \STATE Solve the optimization problem in Equation \ref{eq: optimization} based on the estimated gap $\widehat{\Delta}(t-1)$.
           \IF{$||a||^2_{G^{-1}_{t-1}} \leq \max \{ \frac{\widehat{\Delta}_{min}^2(t-1)}{f_n} , \frac{(\widehat{\Delta}_a^{c_t}(t-1))^2}{f_n} \}, ~ \forall a \in \mathcal{A}$,}
                \STATE \textcolor{blue}{// Exploitation}
                \FOR{Each arm $a \in \mathcal{A}$}
                \STATE $\hat{\mu}_a(t-1) = \max\{ \min\{U_{a,\mathbf{c}}, a^\top \hat{\bm\theta}_{t-1} \}, L_{a,\mathbf{c}}\}$
                \ENDFOR
                \STATE Pull arm $a_t = \argmax_{a \in \mathcal{A}}\hat{\mu}_a(t-1)$.
           \ELSE
           \STATE \textcolor{blue}{//Wasted (LinUCB) Exploration}
           \STATE $s(t) = s(t-1) + 1$
           \IF{$N_a(t-1) \geq \min( T_a(\widehat{\Delta}(t-1)), f_n/(\widehat{\Delta}_{min}(t-1)) )^2, ~ \forall a \in \mathcal{A}$,}
                \STATE  Pull an arm following Equation \ref{eq: estimated ucb}.
           \ELSE
           \STATE Calculate $b_1,b_2$ following Equation \ref{eq: arm exploration}.      
                \IF{$N_{b_2}(t-1) \leq \epsilon_ts(t-1)$}
                \STATE \textcolor{blue}{// Forced Exploration}
                \STATE Pull arm $a_t = b_2$.
                \ELSE 
                \STATE \textcolor{blue}{// Unwasted Exploration}
                \STATE Pull arm $a_t = b_1$.
           \ENDIF
           \ENDIF
           \ENDIF
           \STATE Observe reward and update $\widehat{\bm\theta}_t, \widehat{\Delta}_a^{c_t}(t), \widehat{\Delta}_{min}(t)$
           \ENDFOR
 
	\end{algorithmic}
	\label{alg: oam}
\end{algorithm}

\begin{theorem}[Regret of OAM-PCB]
Given causal bounds $\mathbb{E}[Y_{a,\mathbf{c}}] \in [L_{a,\mathbf{c}}, U_{a,\mathbf{c}}] $ over $a \in \mathcal{A}$, the asymptotic regret of optimal allocation matching policy  augmented with prior causal bounds is bounded by

\begin{equation}
R_{\pi_{\text{oam}}}(T) \leq log(T) \cdot \mathcal{V}(\bm{\theta}, \mathcal{A})    
\end{equation}
where $\mathcal{V}(\bm{\theta}, \mathcal{A})$ denotes the optimal value of the optimization problem defined as:
\begin{equation}
  \inf_{\alpha_{a,\mathbf{c} \in [0,\infty]}} \sum_{c = 1}^{|\mathcal{C}|} \sum_{a: \mathbf{U}_{a,c} \geq \mu_c^*} \alpha_{a,c}\Delta_{a}^c
\label{eq: optimization}
\end{equation}
subject to the constraint that for any context $\mathbf{c}$ and suboptimal arm $a \in \mathcal{A}$,
\begin{equation}
    a^\top  \bigg( \sum_{c=1}^{|\mathcal{C}|} \sum_{a: \U_{a,c} \geq \mu_c^*} \alpha_{a,c} a a^\top \bigg)^{-1} a \leq \frac{(\Delta_a^c)^2}{2}   
\end{equation}
\label{thm: oam_ucb}
\end{theorem}

In Theorem \ref{thm: oam_ucb} $c$ indexes a domain value of the context vector $\BC$, $\mu_c^* = \langle \bm{\theta},a^*_c 
 \rangle$ is the mean reward of the best arm given context $c$, $\Delta_{a}^{\mathbf{c}} = \langle \bm{\theta}, a_c^* - a \rangle$ is the suboptimality gap and $\Delta_{min} = \min_{\mathbf{c} \in [|\mathcal{C}|]}  \min_{a \in \mathcal{A},\Delta_{a}^{\mathbf{c}}>0}\Delta_{a}^{\mathbf{c}}$. We next derive the asymptotic regret bound of OAM-PCB and show our theoretical results. 

  \section{Proof of Theorem \ref{thm: oam_ucb}}
\begin{proof}

We first define $\Delta_{max} = \max_{a,c}\Delta_a^c$ and $\mathcal{A}^-_{c} = \{ a: U_{a,c}\geq \mu^*_c \}$. According to Lemma 3.2 in \cite{hao2020adaptive} $G_t$ is guaranteed to be invertible since the arm set $\mathcal{A}$ is assumed to span $\mathbb{R}^d$. The regret during the initialization is at most $d \Delta_{max} \approx o(log(T))$. We can thus ignore the regret during the initialization phase in the remaining proof.

To prove the regret during the exploration-exploitation phase, we first define the event $\mathcal{B}_t$ as follows:
\begin{equation}
\mathcal{B}_t = \bigg\{  \exists t \geq l, \exists a \in \mathcal{A}, s.t. ~ |a^\top \hat{\bm \theta}_t - a^\top \bm \theta| \geq ||a||_{G^{-1}_t} f_n^{1/2}  \bigg\}
\label{eq: beta}
\end{equation}

By choosing $\delta = 1/T$, from Lemma A.2 in \cite{hao2020adaptive} we have $P(\mathcal{B}_t) \leq 1/T$. We thus decompose the cumulative regret by applying optimal allocation matching (oam) policy with respect to event $\mathcal{B}_t$ as follows:

\begin{equation}
\begin{split}
&R_{\pi_{\text{oam}}}(T) = \mathbb{E}[ \sum_{t=1}^T \sum_{a \in \mathcal{A}} \Delta_{a}^{c_t} \mathds{1}(a_t =a) ] =  \\
&  \mathbb{E}[ \sum_{t=1}^T \sum_{a \in \mathcal{A}} \Delta_{a}^{c_t} \mathds{1}(a_t =a,\mathcal{B}_t) ] +  \mathbb{E}[ \sum_{t=1}^T \sum_{a \in \mathcal{A}} \Delta_{a}^{c_t} \mathds{1}(a_t =a,\mathcal{B}^c_t ) ]    
\end{split}
\label{eq: regret}
\end{equation}

The first term of Equation \ref{eq: regret} could be asymptotically bounded by $o(log(T))$:
\begin{equation}
\begin{split}
&\limsup_{T \to \infty} \frac{ \mathbb{E}[ \sum_{t=1}^T \sum_{a \in \mathcal{A}} \Delta_{a}^{c_t} \mathds{1}(a_t =a,\mathcal{B}_t) ] }{log(T)}  \\
= &\limsup_{T  \to \infty} \frac{ \mathbb{E}[ \sum_{t=1}^T \Delta_{a_t}^{c_t} \mathds{1}(\mathcal{B}_t) ] }{log(T)} \\
\leq & \limsup_{t \to \infty} \frac{ \Delta_{max} \sum_{t=1}^T P(\mathcal{B}_t)  }{log(T)} \leq  \limsup_{T  \to \infty} \frac{ \Delta_{max} \sum_{t=1}^T 1/T  }{log(T)} \\
= &\limsup_{T  \to \infty} \frac{\Delta_{max}}{log(T)} = 0
\end{split}
\label{eq: first term}
\end{equation}

To bound the second term in Equation \ref{eq: regret}, we further define the event $\mathcal{D}_{t,c_t}$ by
\begin{equation}
\begin{split}
&\mathcal{D}_{t,c_t} =\\
&\bigg\{ \forall a \in \mathcal{A}, ||a||^2_{G_t^{-1}} \leq   \max \{ \frac{\widehat{\Delta}_{min}^2(t-1)}{f_n} , \frac{(\widehat{\Delta}_a^{c_t}(t-1))^2}{f_n} \}  \bigg\}  
\end{split}
\end{equation}

At time $t$ the algorithm exploits under event $\mathcal{D}_{t,c_t}$. Under event $\mathcal{D}^c_{t,c_t}
$ the algorithm explores at round $t$. We then further decompose the second term in Equation \ref{eq: regret} into the sum of exploitation regret and exploration regret:

\begin{equation}
\begin{split}
\mathbb{E}[ \sum_{t=1}^T \sum_{a \in \mathcal{A}} \Delta_{a}^{c_t} &\mathds{1}(a_t =a,\mathcal{B}^c_t ) ] \\
=& \underbrace{\mathbb{E}[ \sum_{t=1}^T \sum_{a \in \mathcal{A}} \Delta_{a}^{c_t} \mathds{1}(a_t =a,\mathcal{B}^c_t, \mathcal{D}_{t,c_t}) ]}_\text{exploitation regret}\\
 +&  \underbrace{\mathbb{E}[ \sum_{t=1}^T \sum_{a \in \mathcal{A}} \Delta_{a}^{c_t} \mathds{1}(a_t =a,\mathcal{B}^c_t,\mathcal{D}^c_{t,c_t} ) ] }_\text{exploration regret}  
\end{split} 
\label{eq: further decomposition}
\end{equation}

We then bound those two terms by Lemma \ref{thm: exploitation regret} and Lemma \ref{thm: exploration regret} accordingly.

\begin{lemma}
The exploitation regret satisfies 
\begin{equation}
\limsup_{T  \to \infty} \frac{\mathbb{E}[ \sum_{t=1}^T \sum_{a \in \mathcal{A}} \Delta_{a}^{c_t} \mathds{1}(a_t =a,\mathcal{B}^c_t, \mathcal{D}_{t,c_t}) ]}{log(T)} = 0    
\end{equation}
\label{thm: exploitation regret}
\end{lemma}

\begin{lemma}
The exploration regret satisfies 
\begin{equation}
\limsup_{T  \to \infty} \frac{\mathbb{E}[ \sum_{t=1}^T \sum_{a \in \mathcal{A}} \Delta_{a}^{c_t} \mathds{1}(a_t =a,\mathcal{B}^c_t, \mathcal{D}^c_{t,c_t}) ]}{log(T)} \leq  \mathcal{V}(\bm \theta, \mathcal{A})    
\end{equation}
where $\mathcal{V}(\bm \theta, \mathcal{A})$ is defined in Theorem \ref{thm: oam_ucb}.
\label{thm: exploration regret}
\end{lemma}

Combining the bounds of exploitation and exploration regrets leads to the results below: 
\begin{equation}
    \limsup_{T  \to \infty} \frac{\mathbb{E}[ \sum_{t=1}^T \sum_{a \in \mathcal{A}} \Delta_{a}^{c_t} \mathds{1}(a_t =a,\mathcal{B}^c_t) ]}{log(T)} \leq  \mathcal{V}(\bm \theta, \mathcal{A}) 
\end{equation}
\end{proof}
Finally, combining the results in Equation \ref{eq: first term} leads to the asymptotic regret bound of Algorithm \ref{alg: oam}:

\begin{equation}
R_{\pi_{\text{oam}}}(T) \leq log(T) \cdot \mathcal{V}(\bm{\theta}, \mathcal{A})     
\end{equation}

\subsection{Proof of Lemma \ref{thm: exploitation regret}}
\begin{proof}
Under the event $\beta^c_t$ defined in Equation \ref{eq: beta}, we have
\begin{equation}
    \max_{a \in \mathcal{A}}|\langle \hat{\bm\theta_t} - \bm\theta, a   \rangle| \leq ||a||_{G_t^{-1}}f_n^{1/2}
\end{equation}

We further bound $||a||_{G_t^{-1}}$ under event $\mathcal{D}_{t,c_t}$ by:
\begin{equation}
||a||_{G_t^{-1}} \leq \max \bigg( \frac{\widehat{\Delta}_{min}^2}{f_n} , 
\frac{\widehat{\Delta}_a^c(t)^2}{f_n}\bigg) = \frac{(\widehat{\Delta}_a^c(t))^2}{f_n}
\end{equation}

We further define $\tau_a$ for each $a \in \mathcal{A}$ as
\begin{equation}
\begin{split}
    \tau_a = \min \bigg\{ N: \forall t\geq d, &\mathcal{D}_t,c_t ~ \text{occurs}, N_a(t) \geq N,\\
    &\text{implies} ~ |\langle  \hat{\bm\theta}_t - \bm\theta, a \rangle | \leq  \frac{\Delta_{min}}{2}\bigg\}    
\end{split}
\end{equation}
and decompose the exploitation regret with respect to the event $\{ N_{\hat{a}^*_c(t)}(t)  \geq \tau_{\hat{a}_c^*(t)} \}$  defined in \cite{hao2020adaptive} as follows:
\begin{equation}
\begin{split}
   & \mathbb{E}\bigg[ \sum_{t=1}^T\sum_{a \in \mathcal{A}^-_{c_t}} \Delta_{a}^{c_t} \mathds{1}(a_t =a,\mathcal{B}^c_t, \mathcal{D}_{t,c_t})     \bigg] \\
   \leq & \mathbb{E}\bigg[ \sum_{c=1}^{|\mathcal{C}|}\sum_{t=1}^T\sum_{a \in \mathcal{A}^-_{c}} \Delta_{a}^{c} \mathds{1}(a_t =a,\mathcal{B}^c_t, \mathcal{D}_{t,c}, N_{\hat{a}^*_c(t)}(t)  \geq \tau_{\hat{a}_c^*(t)})  \bigg] \\
   + & \mathbb{E}\bigg[ \sum_{c=1}^{|\mathcal{C}|}\sum_{t=1}^T\sum_{a \in \mathcal{A}^-_{c}} \Delta_{a}^{c} \mathds{1}(a_t =a,\mathcal{B}^c_t, \mathcal{D}_{t,c}, N_{\hat{a}^*_c(t)}(t)  < \tau_{\hat{a}_c^*(t)})  \bigg]
\end{split}
\label{eq: exploration decomposition}
\end{equation}

When $a^*_c = \hat{a}^*_c(t)$ the first term in Equation \ref{eq: exploration decomposition} equals $0$, we next bound the second term:
\begin{equation}
\begin{split}
& \mathbb{E}\bigg[ \sum_{c=1}^{|\mathcal{C}|}\sum_{t=1}^T\sum_{a \in \mathcal{A}^-_{c}} \Delta_{a}^{c} \mathds{1}(a_t =a,\mathcal{B}^c_t, \mathcal{D}_{t,c}, N_{\hat{a}^*_c(t)}(t)  < \tau_{\hat{a}_c^*(t)})  \bigg]   \\
\leq& \mathbb{E}\bigg[ \sum_{c=1}^{|\mathcal{C}|}\sum_{t=1}^T\mathds{1}(a_t =a,\mathcal{B}^c_t, \mathcal{D}_{t,c}, N_{\hat{a}^*_c(t)}(t)  < \tau_{\hat{a}_c^*(t)})  \bigg] \Delta_{max}\\
\leq& \sum_{c=1}^{|\mathcal{C}|}\sum_{a \in \mathcal{A}} \mathbb{E}[\tau_a]\Delta_{max} \leq \sum_{a \in \mathcal{A}}\mathbb{E}(\tau_a)\Delta_{max}
\end{split}
\end{equation}

Combining  the results together leads to the desired results: 
\begin{equation}
\begin{split}
     &\limsup_{T  \to \infty} \frac{\mathbb{E}[ \sum_{t=1}^T \sum_{a \in \mathcal{A}} \Delta_{a}^{c_t} \mathds{1}(a_t =a,\mathcal{B}^c_t, \mathcal{D}_{t,c_t}) ]}{log(T)} \\        
 \leq& \limsup_{T  \to \infty} \frac{|\mathcal{A}|\Delta_{max}(8(1+1/log(n)) + 4cdlog(dlog(n)))}{\Delta^2_{min}log(T)} = 0
\end{split}
\end{equation}
\end{proof}

\subsection{Proof of Lemma \ref{thm: exploration regret}}

\begin{proof}
Let $\mathcal{M}_t$ denote the set that records the index of action sets that has not been fully explored until round $t$:
\begin{equation}
\mathcal{M}_t = \bigg\{ m:  \exists a \in \mathcal{A}^m, N_a(t) \leq \min\{ f_n/\hat{\Delta}^2_{min}(t), T_a(\hat{\Delta}(t))  \}  \bigg\}
\end{equation}

Under the event $\mathcal{D}^c_{t,c_t}$ we have $\mathcal{M}_t \neq \emptyset$. 
We decompose the exploration regret into two terms:  regret under unwasted exploration and wasted exploration, according to whether $c_t$ belongs to $\mathcal{M}_t$.
\begin{equation}
\begin{split}
    \mathbb{E}[ \sum_{t=1}^T \sum_{a \in \mathcal{A}} &\Delta_{a} \mathds{1}(a_t =a,\mathcal{B}^c_t,\mathcal{D}^c_{t,c_t} ) ] = \\
    & \underbrace{\mathbb{E}[ \sum_{t=1}^T \sum_{a \in \mathcal{A}} \Delta_{a} \mathds{1}(a_t =a,\mathcal{B}^c_t,\mathcal{D}^c_{t,c_t}, c_t \in \mathcal{M}_t)]}_\text{unwasted exploration} \\
    +&  \underbrace{\mathbb{E}[ \sum_{t=1}^T \sum_{a \in \mathcal{A}} \Delta_{a} \mathds{1}(a_t =a,\mathcal{B}^c_t,\mathcal{D}^c_{t,c_t}, c_t \notin \mathcal{M}_t)]}_\text{wasted exploration} 
\end{split}
\end{equation}


Following the proof procedure of Lemma B.1 and B.2 in \cite{hao2020adaptive} by substituting with the reduced arm set $\mathcal{A}^-_{c} = \{ a: U_{a,c}\geq \mu^*_c \}$, we show that the regret regarding the wasted explorations is bounded by $o(log(T))$, and the regret regarding to the unwasted explorations is bounded by $log(T) \cdot \mathcal{V}(\bm\theta,\mathcal{A})$. Combing the bounds of these two terms leads to our conclusion.
\end{proof}

\section{Implementation Details of UCB-PCB Algorithm}

Our prior causal bounds can also be incorporated into non-contextual bandits. 
In non-contextual bandit setting, the goal is to calculate $\mathbb{E}[Y|do(\X = \x)]$ for each $\x $ and identify the best arm. Recall Equation 6 in the main text we have $P_{\x}(y|\mathbf{c}) =  \frac{P_{\x}(y,\mathbf{c})}{P_{\x}(\mathbf{c})}$. Simply replacing the outcome variable with $y$ and removing the term $P_{\x}(\mathbf{c})$ from Equation 7 and 10 in the main text leads to the causal bound for the interventional distribution $p_{\x}(y)$. 
We derive the UCB-PCB algorithm, a non-contextual UCB-based multi-arm bandit algorithm enhanced with prior causal bounds shown as follows.
\begin{algorithm}[htp]
	\caption{UCB Algorithm with Prior Causal Bounds (UCB-PCB)}
	\begin{algorithmic}[1]
		\STATE  \textbf{Input:} Time horizon $T$, arm set $\mathcal{A}$, causal Graph $\mathcal{G}$, prior causal bounds $\{[L_{a},  U_{a}]\}_{a \in  \mathcal{A}}$.
		\STATE  \textbf{Initialization:} Values assigned to each arm $a$: $\hat{\mu}_{a}$,  $N_a(0) = 0$.

	    \FOR {$t = 1,2,3,...,T$}
	    \FOR{each arm $a \in \mathcal{A}$}
	       \STATE $UCB_{a}(t-1) = \hat{\mu}_{a}(t-1) + \sqrt{\frac{2log(1/\delta)}{N_{a}(t-1)}}$.
	       \STATE $\widehat{UCB}_{a}(t-1) = \max\{ \min\{U_{a}, UCB_{a}(t-1) \}, L_{a}\}$
	    \ENDFOR
	    \STATE Pull arm $a_t = \argmax_{a \in \mathcal{A}}\widehat{UCB}_{a}(t-1)$
	    \STATE Observe $Y_t$ and update upper confidence bounds accordingly.
	    \ENDFOR
	\end{algorithmic}
	\label{alg:causalucb}
\end{algorithm}

We also derive the regret bound of our UCB-PCB algorithm in Theorem \ref{thm: c-ucb} and demonstrate the potential improvement due to prior causal bounds. The derived instance dependent regret bound indicates the improvement could be significant if we obtain concentrated causal bounds from observational data and consequently exclude more arms whose causal upper bounds are less than $\mu^*$. 

\begin{theorem}[Regret of UCB-PCB algorithm]
Suppose the noise term is 1-subgaussian distributed, let $\delta = 1/T^2$, the cumulative regret for k-arm bandit bounded by:
\[ R(T) = 3 \sum_{a=1}^{k} \Delta_a + \sum_{a: \BoldU_a \geq \mu^*} \frac{16log(T)}{\Delta_a}\]
$\Delta_a$ denotes the reward gap between arm $a$ 
and the optimal arm $a^*$.
\label{thm: c-ucb}
\end{theorem}


\section{Proof of Theorem \ref{thm: c-ucb} }
We first decompose the cumulative regret up to time $T$:
\begin{equation}
    R(T) = \sum_{a=1}^k \Delta_a \mathbb{E}[N_a(T)]
\label{eq: regret_cucb}
\end{equation}
Let $E_a$ be the event defined by:
\[  E_a =  \bigg\{ \mu^* < \min_{t \in [T]} UCB^*(t,\delta) \bigg\} \cap \bigg\{ \hat{\mu}_{au_a} + \sqrt{\frac{2}{u_a}log(\frac{1}{\delta})}  < \mu^*  \bigg\}  \]
where $u_a \in [T]$ is a constant.
Since $N_a(T) \leq T$, we have
\begin{equation}
\begin{split}
\mathbb{E}[N_a(T)] &=  \mathbb{E}[\mathds{1}\{ E_a\}N_a(T)] + \mathbb{E}[\mathds{1}\{ E^c_a\}N_a(T)]\\
&\leq u_a + P(E^c_i)T     
\end{split}   
\label{eq: played times}
\end{equation}
We will show that if $E_a$ occurs, the number of times arm $a$ is played up to time $T$ is upper bounded (Lemma \ref{lemmma: number}), and the complement event $E_a^c$ occurs with low probability (Lemma \ref{lemmma: p(E)}).

\begin{lemma}
If $E_a$ occurs, the times arm $a$ is played is bounded by:
\begin{equation}
\mathbb{E}[N_a(T)] \leq \left\{ 
\begin{aligned}
0 & , & \textit{if}~~ U_a < \mu^* \\
3 + \frac{16log(T)}{\Delta_a^2} & , & \textit{otherwise}
\end{aligned}
\right.
\end{equation}
\label{lemmma: number}
\end{lemma}

\begin{lemma}
   \[ P(E_a^c) \leq T\delta + \text{exp} \bigg(  -\frac{u_ac^2\Delta_a^2}{2} \bigg)\]
\label{lemmma: p(E)}
\end{lemma}

Combining the results of the two lemmas we have 
\begin{equation}
P(E^c_i) \leq T\delta + \text{exp} \bigg( -\frac{u_ac^2\Delta_a^2}{2} \bigg)
\end{equation}

Then by substituting the result from the two lemmas into Equation \ref{eq: played times}, we have 
\begin{equation}
    \mathbb{E}[N_a(T)] \leq u_a + T \bigg( T\delta + \text{exp} \bigg( -\frac{u_ac^2\Delta_a^2}{2} \bigg) \bigg)
\end{equation}

We next aim to choose a suitable value for $u_a \in [T]$. Directly solving Equation \ref{eq: u_a} and taking the minimum   value in the solution space leads to a legit value of $u_a$:
\begin{equation}
u_a = \bigg\lceil\frac{2log(1/\delta)}{(1-c)^2\Delta_{a}^2} \bigg\rceil
\label{eq: u_a value}
\end{equation}

Then we take $\delta = 1/{n^2}$ and $u_a$ with the value in Equation \ref{eq: u_a value} to get the following equation:
\begin{equation}
\begin{split}
\mathbb{E}[N_a(T)] &\leq u_a + 1 + T^{1-2c^2/(1-c)^2}   \\
&= \bigg\lceil \frac{2log(1/\delta)}{(1-c)^2\Delta_{a}^2} \bigg\rceil  + 1 + T^{1-2c^2/(1-c)^2}      
\end{split}
\end{equation}

By substituting $c = 1/2$ in the above equation we obtain
\begin{equation}
\mathbb{E}[N_a(T)] \leq 3 + \frac{16log(T)}{\Delta_{a}^2}    
\label{eq: bound of number}
\end{equation}

Finally, substituting $\mathbb{E}[N_a(T)]$ with the bound above for each arm $a \in \mathcal{A}$ 
 in Equation \ref{eq: regret_cucb} leads to the desired regret bound:
\[ R(T) \leq 3 \sum_{a=1}^{k} \Delta_a + \sum_{a: \BoldU_a \geq \mu^*} \frac{16log(T)}{\Delta_a}\]

\subsection{Proof of Lemma \ref{lemmma: number}}
We derive the proof by contradiction. Suppose $N_a(T) > u_a$, there would exist a round $t \in [T]$ where $N_a(t-1) = u_a$ and $a_t = a$. By the definition of $E_a$ we have 
\begin{equation}
\begin{split}
 UCB_a(t-1,\delta) &= \hat{u}_a(t-1) + \sqrt{\frac{2log(1/\delta)}{N_a(t-1)}} \\
 &= \hat{u}_{au_a} + \sqrt{\frac{2log(1/\delta)}{u_a}} 
 < u^* < UCB^*(t-1,\delta)    
\end{split}
\end{equation}
We thus have $a_t = \argmax_{i}UCB_i(t-1,\delta) \neq a$, which leads to a contradiction. As a result, if $E_a$ occurs we have $N_a(T) \leq u_a$. 

\subsection{Proof of Lemma \ref{lemmma: p(E)}}
According to the definition $E_a^c$ is defined as:
\begin{equation}
     E_a^c = \underbrace{\bigg\{ \mu^* \geq \min_{t \in [T]} UCB^*(t,\delta) \bigg\}}_\text{term 1}  \cup \underbrace{\bigg\{ \hat{\mu}_{au_a} + \sqrt{\frac{2log(1/\delta)}{u_a}}  \geq \mu^*  \bigg\}}_\text{term 2}
\label{eq: ea}
\end{equation}
We decompose term 1 according to the definition of $UCB^*(t,\delta)$:
\begin{equation}
\begin{split}
\bigg\{ \mu^* \geq \min_{t \in [T]} UCB^*(t,\delta) \bigg\} \subset   \bigg\{ \mu^* \geq \min_{s \in [T]} \hat{u}_{1s} + \sqrt{\frac{2log(1/\delta)}{s}} \bigg\}  \\
= \bigcup_{s \in [T]} \bigg\{ \mu^* \geq  \hat{u}_{1s} + \sqrt{\frac{2log(1/\delta)}{s}} \bigg\}    
\end{split}
\end{equation}

We next apply corollary 5.5 in \citep{lattimore2020bandit} and leverage union bound rule of independent random variables to further upper bound term 1 by $n\delta$ as follows: 
\begin{equation}
\begin{split}
&P\bigg(  u^* \geq  \min_{t \in {T}} UCB^*(t,\delta)\bigg) \\
\leq& P\bigg( \bigcup_{s \in [T]} \bigg\{ \mu^* \geq  \hat{u}_{1s} + \sqrt{\frac{2log(1/\delta)}{s}} \bigg\}  \bigg) \\
\leq& \sum_{s=1}^{T}P \bigg(  \mu^* \geq  \hat{u}_{1s} + \sqrt{\frac{2log(1/\delta)}{s}} \bigg) \leq n\delta     
\end{split}
\end{equation}

Next we aim to bound term 2 in Equation \ref{eq: ea}. We proceed by assuming $u_a$ is large enough such that 
\begin{equation}
    \Delta_{a} - \sqrt{\frac{2log(1/\delta)}{u_a}} \geq c \Delta_{a}
\label{eq: u_a}
\end{equation}
for some constant $c \in (0,1)$ to be chosen later. Since $u^* = u_a + \Delta_{a}$, according to corollary 5.5 in \citep{lattimore2020bandit} we have 
\begin{equation}
\begin{split}
&P\bigg(\hat{\mu}_{au_a} + \sqrt{\frac{2log(1/\delta)}{u_a}}  \geq \mu^* \bigg) \\
=& P\bigg(\hat{\mu}_{au_a} - u_a \geq \Delta_{a} - \sqrt{\frac{2log(1/\delta)}{u_a}}   \bigg)   \\
\leq& P \bigg( \hat{\mu}_{au_a} - u_a \geq c\Delta_{a} \bigg) \leq  \text{exp}(-\frac{u_ac^2\Delta^2_a}{2})    
\end{split}
\end{equation}

\section{Experimental Settings}
Table \ref{tab: adult} shows the comparison results in offline evaluation phase.  Biased and CP denote two biased estimation baselines. \textit{lb} and \textit{ub} denote the lower bound and upper bound derived by our BCE algorithm for the conditional causal effect related to a value of the context vector. We report the causal bound and the estimated values among 16 different values of the context vector. Table \ref{table: cp} demonstrates the conditional probabilities for generating the synthetic dataset.

\begin{table}[t]
	\begin{center}
		\caption{Reward Estimation for the Synthetic Data.}
		\scalebox{1}
		{
			\begin{tabular}{|c|c|c|c|c|c|}\hline
			\multirow{2}{*}{\textit{Index}}	 & \multirow{2}{*}{\textbf{CP}} & \multirow{2}{*}{\textbf{Biased}}   & \multicolumn{2}{c|}{\textbf{BCE}} & \multirow{2}{*}{\textbf{Truth}} \\ \cline{4-5}
			    	&           &     & \textit{lb}  & \textit{ub} &  \\ \hline
				$ 1$    &  0.313        & 0.364          & 0.246      & 0.385        &  0.283   \\ \hline
                $ 2$    &  0.336        & 0.351          & 0.236      & 0.371        &  0.278     \\ \hline
                $ 3$    &  0.533        & 0.346          & 0.233      & 0.367        &  0.275  \\ \hline
                $ 4$    &  0.510        & 0.362          & 0.244      & 0.384        &0.283  \\ \hline
                $ 5$    &  0.517        & 0.539          & 0.350      & 0.565        &0.448   \\ \hline
                $ 6$    &  0.494        & 0.519          & 0.376      & 0.545        &0.440   \\ \hline
                $ 7$    &  0.657        & 0.513          & 0.371      & 0.538        &0.436  \\ \hline
                $ 8$    &  0.710        & 0.537          & 0.389      & 0.563        &0.448   \\ \hline
                $ 9$    &  0.377        & 0.492         & 0.296      & 0.505       &0.419  \\ \hline
                $ 10$    & 0.386         & 0.474          & 0.285       & 0.486        &0.411   \\ \hline
                $ 11$    & 0.525         & 0.468          & 0.282      & 0.480        &0.407   \\ \hline
                $ 12$    & 0.518         & 0.490          & 0.295      & 0.503        &0.419 \\ \hline
                $ 13$    & 0.547         & 0.652          & 0.414      & 0.665        &0.580 \\ \hline
                $ 14$    & 0.513         & 0.628          & 0.409      & 0.640        &0.569  \\ \hline
                $ 15$    & 0.756         & 0.620          &  0.404     & 0.632        &0.564 \\ \hline
                $ 16$    & 0.681         & 0.649          & 0.324     & 0.662        &0.580  \\ \hline
		\end{tabular}}
		\label{tab: adult}
	\end{center}
\end{table}

\begin{table*}[ht]
\caption{Conditional Probability Table for Data Generation}
\centering
  \begin{tabular}{|c|c|c|c|c|} \hline
     Variable & Distributions $P(V_i = 1| Pa_{V_i})$\\ \hline
     $U_1$ & $P(U_1 = 1) = 0.4$  \\ \hline
     $U_2$ & $P(U_2 = 1) = 0.6$ \\ \hline
     $X_1$ & $P(X_1 = 1|U_1 = u_1) = (\mathds{1}_{\{u_1=1\}} + 0.5)/2$   \\ \hline
     $X_2$ & $P(X_2 = 1|U_2 = u_2) = (\mathds{1}_{\{u_2 = 1\}}+0.3)/2$   \\ \hline
     $I_1$ & $P(I_1 = 1|X_1 = x_1, C_1 = c_1) = (\mathds{1}_{\{x_1 = 1\}}+ \mathds{1}_{\{c_1=1\}})/4+0.3$  \\ \hline
     $Y$ & \makecell[c]{$P(Y=1|C_1=c_1,U_1=u_1,X_2=x_2,I_1=i_1) = (\mathds{1}_{\{c_1=1\}}+\mathds{1}_{\{u_1 = 1\}}+\mathds{1}_{\{x_2=1\}}+\mathds{1}_{\{i_1 = 1\}})/6 + 0.1$ } \\ \hline
     $C_1$ & $P(C_1 = 1) = 0.5$ \\ \hline
     \multirow{2}{*}{S} & $P(S=1|I_1 = i_1) = 0.8$ if $i_1 =1$  \\ 
      & $P(S=1|I_1 = i_1) = 0.1$ if $i_1 =0$ \\ \hline
  \end{tabular}
  \label{table: cp}
\end{table*}

\end{document}